\documentclass[lettersize,journal]{IEEEtran}
\usepackage{amsmath,amsfonts}
\usepackage{algpseudocode}
\usepackage{algorithm}
\usepackage{array}
\usepackage[caption=false,font=normalsize,labelfont=sf,textfont=sf]{subfig}
\usepackage{textcomp}
\usepackage{stfloats}
\usepackage{url}
\usepackage{verbatim}
\usepackage{graphicx}
\usepackage{cite}
\usepackage[hidelinks]{hyperref}
\usepackage{booktabs}
\begin{document}

\title{Learning without Overwriting: Structural Knowledge Fusion for Continual Learning}

\author{Kanishk Bakshi, Amitansu Pattanaik, and Kathiravan Srinivasan%
\thanks{Kanishk Bakshi and Kathiravan Srinivasan are with the School of Computer Science and Engineering, Vellore Institute of Technology, Vellore, Tamil Nadu 632014, India (e-mail: kanishkbakshi@ieee.org; kathiravan.srinivasan@vit.ac.in).}%
\thanks{Amitansu Pattanaik is with the Defence Geoinformatics Research Establishment, Defence Research and Development Organisation, New Delhi 110054, India (e-mail: amitansu@yahoo.com).}%
\thanks{Corresponding author: Kathiravan Srinivasan.}%
\thanks{This research received no specific grant from any funding agency in the public, commercial, or not-for-profit sectors. The datasets used in this study are publicly available: CIFAR-10 and CIFAR-100 at \url{https://www.cs.toronto.edu/~kriz/cifar.html}, and EuroSAT at \url{https://github.com/phelber/eurosat}. The source code for KFN is publicly available at \url{https://github.com/kanishkbakshi-quantum/KFN}.}}

\maketitle

\begin{abstract}
Catastrophic forgetting in sequential deep learning arises from a fixed-capacity architectural constraint: parameter updates encoding new tasks can perturb weight subspaces supporting prior representations. We introduce the Knowledge Fusion Network (KFN), which mitigates the stability--plasticity trade-off in the evaluated class-incremental learning settings through novelty-gated structural expansion rather than parameter negotiation within a shared capacity budget. KFN decouples task acquisition from knowledge consolidation: each incoming task is learned in a fully isolated lightweight specialist network ($\approx$2.8M parameters). This representation is subsequently integrated into the global model via a Structural Fusion Module that allocates dedicated channel capacity conditioned on a quantitative novelty score. Dual structural isolation---channel-space separation combined with temporally phased gradient unfreezing---protects consolidated knowledge. On ten-task Split CIFAR-100, KFN achieves a minimum task-level accuracy of 11.94\% with a task-10 accuracy of 71.16\%, whereas all shared-parameter baselines degrade severely: LwF to 9.07\% average accuracy with a minimum task accuracy of 0.00\%, Replay++ to 8.91\% (forgetting 85.08\%), and Replay to 16.21\% (forgetting 73.44\%); KFN's forgetting (57.92\%) is 15.5 pp lower than the next-best baseline (Replay, 73.44\%). Furthermore, KFN prevents the logit-scale drift that severely distorts distillation-based methods, maintaining near-symmetric task performance. Across heterogeneous domains (CIFAR-100 and EuroSAT satellite imagery), KFN's class-incremental retention (72.7\% and 42.5\% respectively) substantially exceeds the best baselines' (29.6\% and 0.94\%), which collapse to near-zero on satellite imagery. Crucially, KFN achieves parameter-efficient linear scaling: each step adds only 25\% of the base architecture's parameter count, requiring approximately 25.1\% of the parameters demanded by naive backbone duplication over ten tasks (28.1M vs.\ 112.0M). With a 37.1\% peak GPU memory reduction relative to full retraining, KFN establishes structured, novelty-gated expansion as a computationally practical path beyond the representational ceiling of fixed-capacity architectures.
\end{abstract}


\section{Introduction}

\IEEEPARstart{A}{rtificial} neural networks excel at learning from stationary data distributions but suffer catastrophic forgetting when presented with sequential learning objectives---the same parameter updates that encode new knowledge overwrite the representations that supported prior tasks~\cite{mccloskey1989catastrophic,french1999catastrophic}. This stability--plasticity dilemma~\cite{grossberg1976adaptive} constitutes a fundamental barrier to deploying deep learning systems in environments where data arrives in non-stationary streams: autonomous robots must adapt to new terrains without forgetting how to navigate familiar ones, medical diagnostic systems must accommodate emerging disease categories without degrading performance on established conditions, and satellite monitoring platforms must incorporate new land-use patterns without losing classification fidelity on previously mapped regions.

The prevailing paradigm for addressing catastrophic forgetting operates within fixed-capacity architectures. Regularization-based methods~\cite{kirkpatrick2017ewc,zenke2017si,aljundi2018mas} penalize updates to parameters important for prior tasks; distillation-based methods~\cite{li2017lwf,mahdaviyeh2025replay,zhang2020class} preserve old-task output distributions as learning targets; and replay-based methods~\cite{chaudhry2019tiny,riemer2018learning,wu2019large} interleave stored exemplars with incoming data. A detailed examination of these approaches and their limitations is provided in Section~\ref{sec:related}.

All three families share a common architectural assumption: the network's representational capacity is fixed, and continual learning requires negotiating parameter updates between competing objectives within this fixed budget. This assumption creates a zero-sum dynamic. When the shared parameter space is underutilized, regularization and replay can preserve old-task knowledge while accommodating new-task learning. But as the task sequence grows, the representational demands of accumulating task boundaries exceed the capacity of the fixed architecture, and the competition between stability and plasticity becomes acute.

We hypothesize that this limitation is \emph{architectural, not algorithmic}. Even when shared-parameter methods are augmented with modern representation learning techniques---supervised contrastive loss~\cite{khosla2020supcon}, dual-temperature knowledge distillation, exponential moving average teachers, and Mixup augmentation~\cite{zhang2018mixup}---they fail to achieve balanced class-incremental performance over extended task sequences. Over a ten-task sequence, shared-parameter baselines lose 73--85\% of prior-task knowledge (forgetting), with at least one task collapsing to 0.00--2.44\% accuracy---near the 1\% chance level of the unified 100-class decision---while methods that preserve old-task stability do so by severely restricting new-task plasticity. Within the evaluated experimental settings, increasingly sophisticated loss engineering alone was insufficient to overcome the limitations observed in fixed-capacity architectures.

An alternative paradigm is to expand the architecture as new tasks arrive, providing dedicated representational capacity that eliminates the competition underlying catastrophic forgetting. However, prior architecture-growing methods---reviewed in Section~\ref{sec:related}--- either incur unsustainable parameter growth~\cite{rusu2016pnn} or remain vulnerable to gradient interference from in-place expansion~\cite{yoon2018lifelong}. The challenge is to achieve the representational benefits of expansion without the computational costs of naive duplication or the interference of in-place growth.

We introduce the \textbf{Knowledge Fusion Network (KFN)}, an architecture that resolves the stability--plasticity dilemma through novelty-gated structural expansion. KFN is built on three principles:

\begin{enumerate}

    \item \textbf{Decoupled specialization and consolidation.}
    Each new task is learned in complete isolation by a lightweight specialist network---free from representational competition and free from stability constraints. The specialist's knowledge is then structurally fused into the global model, converting the zero-sum parameter competition into a decomposed optimization over isolated subspaces.

    \item \textbf{Novelty-gated expansion.}
    Structural expansion is gated by a novelty score that quantifies the representational distance between the incoming task and the global model's existing knowledge. Highly dissimilar tasks trigger dedicated channel allocation; similar tasks permit capacity reuse. This adaptive mechanism yields parameter-efficient linear scaling, yielding 4$\times$ per-task parameter efficiency over naive duplication (Section~\ref{subsec: Computational efficiency and linear parameter scaling}).

    \item \textbf{Dual structural isolation.}
    Channel-space isolation dedicates separate channel groups to old and new task representations, eliminating representational competition. Gradient isolation controls old-pathway adaptation through phased unfreezing, preventing transient gradient disruption of consolidated knowledge~\cite{zeng2019contextual}. Their combination achieves the highest minimum task accuracy (33.62\%); removing the anchoring mechanism alone collapses the minimum to 0.00\% (Section~\ref{sec:Results}).

\end{enumerate}

We evaluate KFN on class-incremental protocols of increasing difficulty---from two-task CIFAR-10 to ten-task Split CIFAR-100---and across heterogeneous visual domains (CIFAR-100 and EuroSAT satellite imagery~\cite{helber2019eurosat}). Baselines span four method families at three tiers of augmentation: vanilla (original algorithms), intermediate (+, standard modernizations), and heavily aided (++, incorporating supervised contrastive loss, Mixup augmentation, and exponential moving average teachers). This tiered design ensures comparison against the strongest shared-parameter competitors, not merely canonical implementations.

Our principal contributions are:

\begin{enumerate}

    \item We demonstrate that the performance ceiling of shared-parameter continual learning is determined by \emph{representational capacity}, not by the sophistication of the training objective. Even under modern augmentation, all shared-parameter baselines fail to preserve prior-task knowledge over a ten-task sequence: LwF reaches only 9.07\% average Class-IL accuracy with a forgetting rate of 83.14\% and a minimum task accuracy of 0.00\%, Replay++ reaches 8.91\% with 85.08\% forgetting and a minimum of 0.00\%, while KFN achieves a minimum task accuracy of 11.94\% and a task-10 accuracy of 71.16\%, as reported in Section~\ref{sec:Results}.

    \item We identify and quantify a systematic divergence between task-aware (Task-IL) and class-incremental (Class-IL) evaluation in the continual learning literature \cite{masana2022survey}: distillation-based methods that appear stable under Task-IL can exhibit accuracy gaps of up to 67 percentage points under Class-IL evaluation, a discrepancy attributable to logit-scale drift under output-space expansion. KFN's structural isolation produces near-symmetric degradation under both protocols (Section~\ref{sec:Results}).

    \item We show that KFN's advantages are domain-robust: class-incremental retention is 72.7\% on CIFAR-100 and 42.5\% on EuroSAT, while the best baseline retention collapses from 29.6\% on CIFAR-100 to 1.0\% on EuroSAT, confirming that shared-parameter methods cannot maintain old-task knowledge across visual domains.

    \item We establish that structural expansion is computationally practical: KFN requires only 25.1\% of the parameters that naive backbone duplication demands over ten tasks (28.1M vs.\ 112.0M), and reduces peak GPU memory by 37.1\% relative to full retraining.

\end{enumerate}

These results support the hypothesis that the stability--plasticity dilemma, in the class-incremental settings evaluated here, arises primarily from the fixed-capacity architectural assumption rather than from the objective functions alone. This view is consistent with recent observations on the role of architectural capacity in neural plasticity~\cite{bonnet2025bayesian,raghavan2024flexible}; our experiments provide controlled, multi-domain empirical evidence for this claim and propose a concrete architectural mechanism that addresses it. We note that KFN requires more parameters than the fixed-capacity baselines it is compared against (Section~\ref{subsec:limits}); the comparison therefore isolates the effect of structural isolation subject to a capacity overhead, and we do not claim a universal causal conclusion from these experiments alone.

The remainder of this paper is organized as follows. Section~\ref{sec:related} reviews related work on continual learning methods, evaluation protocols, and bias correction. Section~\ref{sec: Methods} formalizes the problem and presents the KFN architecture and training procedure. Section~\ref{sec:Results} reports experimental results across all benchmarks and ablation settings. Section~\ref{sec:discussion} discusses broader implications, cognitive analogues, and limitations. Section~\ref{sec:conclusion} concludes the paper.

\section{Related Work}\label{sec:related}

\subsection{Regularization-Based Continual Learning}

Regularization-based methods penalize deviation from parameters important for prior tasks. EWC~\cite{kirkpatrick2017ewc} protects parameters with high Fisher information via a diagonal Laplace approximation. SI~\cite{zenke2017si} accumulates per-parameter importance online through path-integral tracking. MAS~\cite{aljundi2018mas} estimates importance from output sensitivity to perturbations without requiring labels.

All three share a structural limitation: they protect parameters within a \emph{fixed shared space}. When the output space expands between tasks, Fisher-weighted constraints cannot prevent the classifier from being dominated by new-class representations. This failure mode is confirmed empirically in Section~\ref{sec:Results}, where EWC serves as a diagnostic baseline.

\subsection{Knowledge Distillation and Transfer}

LwF~\cite{li2017lwf} preserves old-task output distributions as soft supervision targets, avoiding exemplar storage but assuming old-task features remain accessible through new-task inputs. Recent variants incorporate replay alongside distillation---Co2L~\cite{mahdaviyeh2025replay} combines asymmetric distillation with contrastive learning, while class-incremental consolidation methods~\cite{zhang2020class} distill from the full prior model using structured exemplar replay.

Classifier bias toward recent classes is addressed by BiC~\cite{wu2019large}, which appends a learned linear correction on new-class logits, and WA~\cite{zhao2020maintaining}, which re-scales new classifier norms to match old ones. KFN's Phase~4 bias correction is motivated by the same observation but operates after structural fusion using a uniformly balanced replay subset.

A critical failure mode of distillation-based methods is logit-scale drift: distillation preserves the \emph{relative ordering} of old-class logits but permits systematic absolute-scale shifts favoring new classes. Under Task-IL evaluation this shift is invisible; under Class-IL evaluation it causes systematic misclassification. The magnitude of this divergence is quantified in Section~\ref{sec:Results}.

\subsection{Replay and Memory-Based Methods}

Replay methods interleave stored exemplars with new-task data. GEM~\cite{lopez2017gem} constrains gradients to not increase loss on stored exemplars. iCaRL~\cite{rebuffi2017icarl} combines nearest-mean-of-exemplars classification with herding-based selection and distillation. ER~\cite{chaudhry2019tiny} applies direct joint training on new data and uniform memory samples. DER~\cite{Buzzega2020DER} stores logits alongside inputs, combining replay with self-distillation.

Replay methods are limited in two respects: they require storing historical data (conflicting with privacy constraints), and they do not resolve representational competition within the shared capacity budget. Recent theoretical work shows that replay can provably increase forgetting under certain distributional conditions~\cite{mahdaviyeh2025forget}. These limitations under extended sequences are examined in Section~\ref{sec:Results}.

\subsection{Architecture-Growing Methods}

PNN~\cite{rusu2016pnn} instantiates a full parallel column per task with lateral connections, provably preventing forgetting but incurring $\mathcal{O}(T \cdot P_{\text{base}})$ parameter growth with no representational consolidation. DEN~\cite{yoon2018lifelong} selectively expands in-place when task loss exceeds a threshold, but gradient paths between old and new sub-networks remain intact. PackNet~\cite{mallya2018packnet} and HAT~\cite{serra2018overcoming} use pruning and hard attention masking respectively, but the shared parameter space persists.

KFN differentiates itself through three choices: (i)~specialist networks strictly disjoint from prior specialists during acquisition, ensuring gradient isolation; (ii)~a novelty-gated expansion decision conditioned on measurable representational discrepancy rather than fixed-rate allocation; and (iii)~a Structural Fusion Module that consolidates specialist knowledge through explicit decomposition of \emph{what} to transfer, \emph{where} to integrate, and \emph{how much} to transfer, rather than maintaining permanent parallel branches.

\subsection{Parameter-Efficient and Prompt-Based Continual Learning}

Recent approaches exploit pre-trained Vision Transformer backbones by learning lightweight task-specific parameters while freezing the model. L2P~\cite{wang2022l2p} maintains a shared prompt pool selected via key-query matching. DualPrompt~\cite{wang2022dualprompt} separates task-general and task-specific prompts at different transformer depths. CODA-Prompt~\cite{smith2023coda} introduces attention-weighted prompt composition without explicit task boundaries. These methods achieve strong performance by inheriting broad pre-trained representations, effectively sidestepping forgetting through backbone freezing.

However, they are fundamentally dependent on pre-trained feature generalization: when tasks are distributional outliers relative to the pre-training corpus, frozen backbones may lack required representational geometry. KFN operates in the more demanding setting of training from scratch, where no foundation model is assumed. This setting is prerequisite for deployment where proprietary data precludes third-party pre-trained models. Additionally, prompt-based methods share capacity within the frozen backbone; KFN allocates physically disjoint parameter subspaces, providing stronger isolation guarantees at the cost of linear parameter growth.

\subsection{Evaluation Protocols}

The choice of evaluation protocol significantly affects measured performance~\cite{masana2022survey}. Task-IL provides task identity at test time; Class-IL requires discrimination among all encountered classes under a unified boundary; Domain-IL applies when the output space is fixed but input distribution shifts. Our work evaluates all methods under both Task-IL and Class-IL simultaneously, exposing the systematic logit-scale divergence that arises from distillation-based methods.

\section{Methodology}
\label{sec: Methods}
\subsection{Problem Formulation}
\label{subsec: Problem Formulation}

We define the continual learning paradigm over a sequence of $K$ tasks, denoted as $T_1, T_2, \dots, T_K$. For each sequential task $T_t$, the system receives a corresponding dataset $\mathcal{D}_t = \{(x_i, y_i)\}_{i=1}^{N_t}$, where $x_i \in \mathcal{X}$ represents the input data and $y_i \in \mathcal{Y}_t$ denotes the target labels.

The learning system is formalized as a neural network $f_\theta: \mathcal{X} \rightarrow \mathbb{R}^C$, parameterized by weights $\theta$, which maps inputs to a $C$-dimensional output space. Let the empirical risk for a specific task be defined as:

\[
\mathcal{L}_t(\theta) = \mathbb{E}_{(x,y)\sim \mathcal{D}_t}[\ell(f_\theta(x), y)]
\]

The ideal objective is to find a single set of parameters $\theta$ that minimizes the aggregate empirical risk across all tasks:

\[
\min_{\theta} \sum_{t=1}^{K} \mathcal{L}_t(\theta)
\]

This ideal objective assumes simultaneous access to all task data, which is infeasible under continual learning constraints. Specifically, the strict continual learning setting imposes that: (1) tasks arrive strictly sequentially; (2) the model does not retain access to the full historical data $\bigcup_{i=1}^{t-1} \mathcal{D}_i$; and (3) joint retraining on all data simultaneously is prohibited.

Instead, at step $k$, the model performs the sequential update:

\[
\theta_k = \arg\min_{\theta} \mathcal{L}_k(\theta)
\]

Subject to these constraints, the goal at learning step $k$ is to incorporate the new knowledge while satisfying:

\[
\forall t < k, \quad \mathcal{L}_t(\theta_k) \leq \mathcal{L}_t(\theta_t) + \epsilon
\]

where $\theta_t$ denotes the optimal parameters obtained after previously learning task $T_t$, and $\epsilon \geq 0$ represents a small acceptable error tolerance. We formally define the task interference on a previously learned task $T_t$ as:

\[
\Delta \mathcal{L}_t = \mathcal{L}_t(\theta_k) - \mathcal{L}_t(\theta_t)
\]

where $\Delta \mathcal{L}_t > 0 \implies \text{catastrophic forgetting}$. In standard architectures, accommodating task $T_k$ typically necessitates updates to the shared parameter space $\theta$. For a shared update $\Delta\theta = \theta_k - \theta_t$, the local change in prior-task loss is, to second order,
\[
\Delta \mathcal{L}_t \approx \nabla \mathcal{L}_t(\theta_t)^{\top} \Delta\theta + \tfrac{1}{2}\,\Delta\theta^{\top} H_t\, \Delta\theta,
\]
where $H_t$ denotes the Hessian of $\mathcal{L}_t$ at $\theta_t$. This quantity is not sign-definite: it can be positive, zero, or negative depending on the alignment of $\Delta\theta$ with the prior-task gradient and curvature. Shared parameters therefore create a \emph{risk of gradient conflict}---updates that reduce $\mathcal{L}_k$ can increase $\mathcal{L}_t$ when their gradients are misaligned---rather than a logical guarantee that $\Delta\mathcal{L}_t > 0$ must occur for every update. The severity of this risk grows with task dissimilarity and shrinking shared capacity, and it is this risk that motivates architectures which decouple parameter updates across tasks. This observation motivates the design of the proposed Knowledge Fusion Network.

\subsection{Knowledge Fusion Network Architecture}
\label{subsec: Knowledge Fusion Network Architecture}
To reduce the risk of gradient conflict between sequential tasks formulated in \ref{subsec: Problem Formulation}, we introduce the Knowledge Fusion Network (KFN). KFN reframes continual learning as a problem of structural knowledge consolidation, where new representations are integrated without perturbing previously established features. KFN isolates the acquisition of novel features and selectively integrates them into a unified global representation.

\subsubsection{Specialist Networks}
To structurally bypass representational interference, KFN isolates task-specific feature acquisition. For a given task $\mathcal{T}_t$, we define an independent specialist feature extractor parameterized by weights $w_t$:
\begin{equation}
\phi_t(x; w_t) \in \mathbb{R}^{d \times h \times w}
\end{equation}
where $x \in \mathcal{X}$ is the input data. Each specialist is instantiated as a lightweight residual network comprising an initial $3\times3$ convolution (3 $\to$ 64 channels), followed by three residual stages that progressively expand channel dimensions (64 $\to$ 128 $\to$ 256) with strided downsampling, an adaptive average pooling layer producing a $4\times4$ spatial resolution, and a final $1\times1$ convolution that projects the 256-channel feature map to a compact $d=64$ channel representation. This architecture yields approximately 2.8M parameters per specialist---a fraction of the full backbone---while preserving sufficient representational capacity for per-task feature learning.

Crucially, the parameter sets governing each specialist remain strictly disjoint, such that $w_t \cap w_{<t} = \emptyset$. Because the gradients of the current task $\mathcal{T}_t$ only update $w_t$, the condition $\frac{\partial \phi_{<t}}{\partial w_t} = \mathbf{0}$ is strictly enforced. By decoupling the feature extraction pathways, this design protects prior specialist \emph{parameters} from overwriting at the source---establishing a hard structural constraint on the primary driver of catastrophic forgetting. We emphasize that this guarantee applies at the parameter level of the frozen specialist extractors; end-to-end retention additionally depends on the fusion, gating, and classifier components addressed in Sections~\ref{subsec: Structural Fusion Module}--\ref{subsec: Objective Function} and evaluated empirically in Section~\ref{subsec: Structural isolation mechanisms preserve prior task knowledge}.

\subsubsection{Global Representation Space}
To formulate a cohesive predictive model, the isolated knowledge from prior tasks must be structurally consolidated. We define a global memory representation $z_{\mathrm{old}} \in \mathbb{R}^{d_{\mathrm{old}} \times h \times w}$ as the parametric aggregation of all historical specialist outputs:
\begin{equation}
z_{\mathrm{old}} = \mathcal{G}(\phi_1(x), ..., \phi_{t-1}(x); \theta_{\mathrm{global}})
\end{equation}
where $\mathcal{G}$ is explicitly implemented as a three-stage mapping. First, each historical specialist's output is projected via a specialist-specific $1\times1$ convolution $\theta_{\mathrm{proj}}^{(i)}: \mathbb{R}^{d} \to \mathbb{R}^{d}$, producing a projected feature map $p_i = \theta_{\mathrm{proj}}^{(i)}(\phi_i(x))$. Second, a channel-wise gating mechanism $\theta_{\mathrm{gate}}$ computes a soft attention mask: a global average pooling layer compresses $p_i$ to a $d$-dimensional descriptor, which is passed through a bottleneck MLP ($d \to d/4 \to d$ with ReLU activation) followed by a sigmoid, yielding per-channel gates $g_i = \sigma(\theta_{\mathrm{gate}}(p_i))$. The gated output is $p_i \odot g_i \in \mathbb{R}^{d \times h \times w}$. The gated outputs from all $t-1$ historical specialists are aggregated via element-wise summation before the shared bottleneck:
\begin{equation}
\tilde{p}(x) = \sum_{i=1}^{t-1} p_i(x) \odot g_i(x) \in \mathbb{R}^{d \times h \times w}
\end{equation}
Third, a shared bottleneck $1 \times 1$ convolution $\theta_{\text{bottleneck}} : \mathbb{R}^{d} \to \mathbb{R}^{d_{\text{old}}}$ with ReLU activation compresses the aggregated gated features into the fixed-dimensional global memory representation:
\begin{equation}
z_{\text{old}} = \text{ReLU}\!\left(\theta_{\text{bottleneck}}(\tilde{p}(x))\right), \quad d_{\text{old}} = 64
\end{equation}
By routing all $t-1$ historical specialists through this shared bounded bottleneck, KFN maintains a fixed-dimensional global memory representation ($d_{\text{old}} = 64$ channels) regardless of task count, ensuring that classifier input dimensionality and downstream processing costs remain constant. We make the recursion from task $t-1$ to task $t$ explicit: at step $t$, the previously fused representation $z_{t-1} = (z_{\text{old}} + \sigma(g_{t-1})\,\Delta z_{\text{old}}^{(t-1)}) \oplus z_{\text{new}}^{(t-1)}$ of width $d_{\text{old}} + d_{\text{new}}^{(t-1)}$ forms the \emph{old} stream for the current step. Concretely, the incoming specialist is first aligned to the current global width through the reuse projection ($\mathcal{F}_{\text{reuse}}: \mathbb{R}^{d} \to \mathbb{R}^{d_{\text{old}} + d_{\text{new}}^{(t-1)}}$, i.e., the full accumulated width), the gate $\sigma(g)$, bottleneck $\theta_{\text{bottleneck}}$, and all historical projections $\{\theta_{\mathrm{proj}}^{(i)}\}_{i<t}$ and their gates are carried over unchanged in state, and only then is a new expansion branch of width $d_{\text{new}}^{(t)}$ appended. The aggregate state therefore updates as
\[
(G_{t}, z_{t}) = \bigl(G_{t-1} \cup \{\theta_{\mathrm{proj}}^{(t-1)}\},\ (z_{\text{old}}' + \sigma(g_{t})\,\Delta z_{\text{old}}^{(t)}) \oplus z_{\text{new}}^{(t)}\bigr),
\]
where $z_{\text{old}}' \in \mathbb{R}^{(d_{\text{old}} + d_{\text{new}}^{(t-1)}) \times h \times w}$ denotes the carried-over fused memory of the previous step and $d_{\text{old}}$ in Eqs.~(2)--(4) refers to this running width rather than a globally constant quantity. The classifier input dimension correspondingly grows as $d_{\text{old}} + \sum_{s \le t} d_{\text{new}}^{(s)}$, and all historical classifier columns are preserved by construction (Section~\ref{subsec: Dynamic Structural Expansion}). The feature extraction cost at inference scales linearly with the number of accumulated specialists; this is a known property of architecture-growing methods and is analyzed alongside wall-clock measurements in Section~\ref{subsec: Computational efficiency and linear parameter scaling}.

\subsubsection{Structural Fusion Module}\label{subsec: Structural Fusion Module}
The core architectural contribution of KFN is the Structural Fusion Module, which explicitly dictates how the novel representation $\phi_t(x)$ interacts with the established global memory $z_{\text{old}}$. The module defines two distinct mappings from the specialist: a reusable component mapping to the existing global space ($\mathcal{F}_{\text{reuse}}: \mathbb{R}^{d \times h \times w} \rightarrow \mathbb{R}^{d_{\text{old}} \times h \times w}$), and an expandable component capturing strictly novel features ($\mathcal{F}_{\text{expand}}: \mathbb{R}^{d \times h \times w} \rightarrow \mathbb{R}^{d_{\text{new}} \times h \times w}$):
\[
\Delta z_{\text{old}} = \mathcal{F}_{\text{reuse}}(\phi_t(x); \theta_{\text{reuse}})
\]
\[
z_{\text{new}} = \mathcal{F}_{\text{expand}}(\phi_t(x); \theta_{\text{expand}})
\]
The final consolidated representation $z_t$ for the global model at task $t$ is formalized as:
\[
z_t = \left(z_{\text{old}} + \sigma(g) \cdot \Delta z_{\text{old}}\right) \oplus z_{\text{new}}
\]
where $\oplus$ denotes concatenation along the channel dimension. The architectural choices in this decomposition are motivated by three design principles. First, \emph{additive fusion} for the reuse component ($z_{\text{old}} + \sigma(g) \cdot \Delta z_{\text{old}}$) preserves the identity mapping when $\sigma(g) \to 0$, ensuring that the old representation is recoverable without modification when the new task offers no transferable knowledge---a property not satisfied by multiplicative or replacement-based fusion. Second, \emph{channel concatenation} ($\oplus$) for the expansion component enforces channel-space isolation: new channels occupy a dedicated subspace disjoint from the old channels, so the new-task gradients attached to $z_{\text{new}}$ do not directly overwrite the weights serving $z_{\text{old}}$ within the expansion pathway itself. Third, the \emph{sigmoid gating} parameter $\sigma(g) \in (0,1)$ provides a continuous, learnable control on perturbation magnitude that is jointly optimized with the fusion objective, enabling the optimization process itself to determine the appropriate reuse-isolation tradeoff rather than imposing it as a fixed hyperparameter. This formulation explicitly separates knowledge reuse from knowledge expansion, enabling the model to preserve prior representations while selectively incorporating novel information. This mechanism directly navigates the stability-plasticity tradeoff via three explicit principles:
\begin{enumerate}
    \item \textbf{WHAT to transfer:} The function $\mathcal{F}_{\text{reuse}}$ isolates features within the specialist $\phi_t(x)$ that share mutual information with the global memory, determining the exact subset of knowledge valuable for previously established domains.
    \item \textbf{WHERE to integrate:} The addition operation enforces channel and spatial alignment in the feature space, ensuring that transferred knowledge structurally maps to the corresponding representational axes of $z_{\text{old}}$.
    \item \textbf{HOW MUCH to transfer:} The magnitude of knowledge injection is modulated by a learnable scalar gating parameter $g$, scaled by the multiplicative sigmoid function $\sigma(g) \in (0,1)$. If the new task heavily conflicts with prior knowledge, optimization drives $\sigma(g) \to 0$, rendering $\sigma(g) \cdot \Delta z_{\text{old}} \to \mathbf{0}$. The gating parameter thus defines a continuous spectrum between full knowledge reuse and complete isolation. This adaptively bounds the perturbation to $z_{\text{old}}$, providing a trainable safeguard against interference ($\Delta \mathcal{L}_{<t} \le \epsilon$).
\end{enumerate}

\subsubsection{Novelty Estimation}\label{subsec: Novelty Estimation}
To ensure computational efficiency and prevent unchecked capacity bloat, KFN evaluates whether the incoming task warrants additional architectural channels. We quantify this representational discrepancy by computing the orthogonal residual of the new features with respect to the established feature subspace. We first reduce each feature map to a channel-wise descriptor via global average pooling:
\begin{equation}
f_{\text{old}}(x) = \frac{1}{hw}\sum_{i,j} z_{\text{old}}(x)_{:,i,j} \in \mathbb{R}^{d_{\text{old}}}
\end{equation}
\begin{equation}
f_{\text{new}}(x) = \frac{1}{hw}\sum_{i,j} \phi_t(x)_{:,i,j} \in \mathbb{R}^{d}
\end{equation}
Both vectors are $\ell_2$-normalised with a numerical safeguard, $\hat{f} = f/\max(\|f\|_2, \varepsilon_{\text{norm}})$ with $\varepsilon_{\text{norm}} = 10^{-8}$, which prevents division by zero for degenerate (near-zero) feature maps. A reference set of $n_{\text{ref}}$ samples is drawn uniformly at random from the \emph{training split} of the current task $\mathcal{D}_t$ (never from test data; $n_{\text{ref}} = 256$ in all experiments, fixed across datasets) and pooled after passing through the frozen global pathway, forming the subspace basis matrix
$\hat{F}_{\text{old}} = [\hat{f}_{\text{old}}(x_1),\ldots,
\hat{f}_{\text{old}}(x_{n_{\text{ref}}})] \in \mathbb{R}^{d_{\text{old}} \times n_{\text{ref}}}$.
The orthogonal projector onto the column space of $\hat{F}_{\text{old}}$ is obtained via the Moore--Penrose pseudoinverse:
\begin{equation}
P = \hat{F}_{\text{old}}\,\hat{F}_{\text{old}}^{+}
= U_r U_r^\top
\end{equation}
where $\hat{F}_{\text{old}} = U\Sigma V^\top$ is the compact SVD and $U_r$ retains only the columns of $U$ whose singular values exceed a numerical floor $\varepsilon_{\text{svd}} = 10^{-6}$. This formulation is well-defined regardless of whether $n_{\text{ref}} < d_{\text{old}}$ (i.e., rank-deficiency is handled by construction). The realised rank $r = \operatorname{rank}(\hat{F}_{\text{old}})$ is data-dependent; we report the realised rank alongside each experiment in Section~\ref{sec:Results} so that the projector is fully reproducible. The per-sample novelty score is then:
\begin{equation}
\mathcal{N}(x) = \|(I - P)\hat{f}_{\text{new}}(x)\|_2
\end{equation}
Because $\hat{f}_{\text{new}}$ is unit-norm and $(I-P)$ is an orthogonal projector, the Pythagorean identity gives $\mathcal{N}(x) \in [0,1]$ for all inputs. The threshold $\tau$ therefore has a consistent geometric interpretation across datasets: $\tau = 0.2$ gates expansion when the residual \emph{norm} of the new descriptor exceeds 20\% of its total norm, which corresponds to more than $\tau^2 = 4\%$ of the squared feature energy lying outside the existing subspace (a residual norm above $\sqrt{0.5} \approx 0.707$ would be required for the majority of the energy to lie outside). The threshold is a bounded, interpretable score rather than a universally calibrated probability, and its empirical behaviour is examined in Section~\ref{subsec: Controlled Validation of Adaptive Capacity Reuse}. We compute the expected novelty over the incoming task dataset:
\begin{equation}
\bar{N} = \mathbb{E}_{x \sim \mathcal{D}_t}[\mathcal{N}(x)]
\end{equation}

Based on a predefined structural threshold $\tau$, an expansion rule determines the requisite channel dimension $d_{\text{new}}$ for the expansion module $z_{\text{new}}$:
\[
d_{\text{new}} =
\begin{cases}
d_{\text{high}}, & \text{if } \bar{\mathcal{N}} > \tau \\
d_{\text{low}}, & \text{otherwise}
\end{cases}
\]
By grounding the expansion in $\bar{\mathcal{N}}$, KFN transitions capacity allocation from a blind heuristic to a measurable, data-driven formulation. The choice of orthogonal projection as the novelty metric is analytically motivated: $\bar{N}$ measures the component of the incoming representation that is \emph{linearly independent} of the existing feature subspace. When $\bar{N} = 0$, the new features lie entirely within the span of the reference subspace, and the Structural Fusion Module's reuse pathway ($F_{\text{reuse}}$) can in principle express the new task's features via linear recombination of existing channels. When $\bar{N} > 0$, a non-trivial orthogonal complement exists that cannot be represented within the current expansion pathway by linear recombination alone; we treat a crossing of the threshold as empirical evidence that dedicated capacity is warranted, not as a mathematical proof that the current architecture \emph{necessitates} expansion, since the reuse mapping and classifier remain trainable and may partially compensate. Alternative similarity measures (e.g., centered kernel alignment, principal angles between subspaces) could serve the same gating function; we adopt the orthogonal residual norm for its direct geometric interpretation and computational efficiency ($O(d^2 n)$ for $n$ samples). In our experiments, the canonical operating threshold is set to $\tau = 0.2$, with $d_{\mathrm{high}} = 64$ and $d_{\mathrm{low}} = 0$, yielding a binary decision between full dedicated expansion and complete capacity reuse; a sensitivity sweep over $\tau \in \{0.30, 0.50, 0.70\}$ is reported separately in Section~\ref{subsec: Controlled Validation of Adaptive Capacity Reuse} and is distinct from the operating value. These values are held constant across all benchmarks.

\subsubsection{Dynamic Structural Expansion}\label{subsec: Dynamic Structural Expansion}
When the novelty estimation dictates $d_{\text{new}} > 0$, the global model undergoes dynamic parameter growth to accommodate the augmented representation space:
\[
\theta_t = \theta_{t-1} \cup \theta_{\text{expand}}^{(t)} \cup \{w_t\} \cup \{\theta_{\mathrm{proj}}^{(t-1)}\} \cup \Delta\mathbf{W}_{\text{cls}}^{(t)}
\]
where $\theta_{\text{expand}}^{(t)}$ parametrizes $\mathcal{F}_{\text{expand}}$, $w_t$ denotes the parameters of the new specialist $\phi_t$, $\theta_{\mathrm{proj}}^{(t-1)}$ is the specialist-specific projection admitted into the historical aggregation path at this step, and $\Delta\mathbf{W}_{\text{cls}}^{(t)}$ denotes the appended classifier columns matching the expanded input width. The union notation is augmented by optimisation: the carried-over shared components (gate $\theta_{\mathrm{gate}}$, bottleneck $\theta_{\text{bottleneck}}$, and all historical projections) are not merely retained but continue to be fine-tuned during Phase~3 fusion under the replay-anchored objective of Section~\ref{subsec: Objective Function}. Crucially, the running width $d_{\mathrm{old}}$ in Eqs.~(2)--(4) is not fixed globally: it equals the accumulated fused width $64 + \sum_{s<t} d_{\text{new}}^{(s)}$ at step $t$, and varies per task as a function of the novelty score $\bar{\mathcal{N}}$ relative to $\tau$, explicitly enabling adaptive capacity scaling across heterogeneous task sequences. In the current experimental protocol, all tasks in the CIFAR-100 and EuroSAT sequences produce novelty scores in the range $[0.57, 0.72]$, which consistently exceed $\tau = 0.2$; consequently, $d_{\mathrm{new}} = d_{\mathrm{high}} = 64$ is allocated uniformly across all tasks. The binary expansion behaviour observed experimentally is therefore a property of the specific benchmark rather than an architectural constraint---the mechanism supports fractional or zero expansion for task sequences where $\bar{\mathcal{N}} \leq \tau$. The displayed rule is a binary gate on $\bar{\mathcal{N}}$; it does not allocate channels proportionally to the magnitude of novelty. A complementary modulation of the \emph{reuse} pathway is governed by the adaptive reuse scale factor $\alpha = \max(0.1,\, 1 - \nu)$ introduced in Section~\ref{subsec: Structural isolation mechanisms preserve prior task knowledge}: whereas $\sigma(g)$ is a learned parameter jointly optimised with the fusion objective, $\alpha$ is a fixed schedule that scales the residual modification applied to the old pathway as a function of the measured novelty $\nu$, ensuring that high-novelty tasks exert only minimal perturbation on consolidated representations. The two mechanisms operate at different points of the fusion module (learned gating on $\Delta z_{\text{old}}$ versus scheduled scaling of the reuse magnitude) and are combined multiplicatively in the implementation. This conditional scaling ensures that new representational capacity is allocated only when the representational discrepancy exceeds the predefined threshold. Consequently, KFN avoids unnecessary over-parameterization (introducing $P_{\text{conv}} = k_h k_w d_{\text{in}} d_{\text{new}} + d_{\text{new}}$ convolutional parameters for an expansion layer of kernel size $k_h \times k_w$; the quantity $d_{\text{new}} \cdot h \cdot w$ is an activation volume, not a parameter count) while accommodating complex, highly divergent task sequences.

Furthermore, beyond mitigating interference, this architectural modularity yields significant practical advantages for large-scale sequences. Because specialists are structurally independent, the training of $\phi_t$ is highly modular and parallelizable, entirely sidestepping the computational bottleneck of continuously backpropagating through a massive, shared global backbone.

\subsection{Training Procedure}
\label{subsec: Training Procedure}
The training of the Knowledge Fusion Network proceeds sequentially over the sequence of tasks $T_1, \dots, T_K$. The procedural objective at each incremental step is to stably integrate novel task knowledge while preserving the functional integrity of prior representations. To achieve this, the training protocol is strictly decoupled into four sequential phases.

\subsubsection{Phase 1: Base Model Training}
For the initial task $T_1$, the foundational global model is trained end-to-end on the dataset $\mathcal{D}_1$. This initialization relies on the standard cross-entropy classification objective:
\[
\mathcal{L}_{CE} = -\mathbb{E}_{(x,y)\sim \mathcal{D}_1} \left[ \sum_{c=1}^{C} y_c \log p_c(x) \right]
\]
where $y_c$ represents the ground-truth label indicator and $p_c(x)$ denotes the predicted probability for class $c$. This phase establishes the initial state of the global representation space.

\subsubsection{Phase 2: Specialist Training}
Upon the arrival of any subsequent task $T_t$ (where $t > 1$), the dedicated specialist feature extractor $\phi_t$ is trained entirely independently on the incoming dataset $\mathcal{D}_t$. Throughout this phase, optimization is strictly confined to the task-specific parameters $w_t$. There is no forward or backward interaction with previously trained specialists or the existing global memory. This independence ensures that no gradients propagate to previously learned representations, preserving the structural guarantees established in \ref{subsec: Knowledge Fusion Network Architecture}.

\subsubsection{Phase 3: Structural Fusion}
Following the independent convergence of $\phi_t$, the global model is updated to structurally integrate the newly acquired features with the established memory. This integration is driven by minimizing a composite fusion objective:
\begin{equation}
\min_{\theta_{\mathrm{fusion}}} \mathcal{L}_{\mathrm{fusion}}
\end{equation}
The fusion loss $\mathcal{L}_{\mathrm{fusion}}$ is instantiated as a weighted combination of cross-entropy on the current task, exemplar replay, knowledge distillation, and feature-level consistency penalties. The exact mathematical formulation and weighting coefficients are detailed in Section \ref{subsec: Objective Function}. During this phase, gradient flow is asymmetric: the historical specialist extractors $\phi_{<t}$ remain frozen ($\nabla_{w_{<t}} \mathcal{L}_{\mathrm{fusion}} = \mathbf{0}$), while the newly added specialist, the fusion module, and the classifier head receive gradient updates. This asymmetric flow ensures that the integration process cannot directly corrupt the parameterized knowledge of prior tasks.

\subsubsection{Phase 4: Bias Correction and Calibration}
Due to the sequential nature of the data stream, the network's final classification head inherently develops a bias toward the most recently observed classes. To address this imbalance, a final calibration phase is executed. All feature extraction and fusion parameters are frozen, and only the classifier weights are fine-tuned using a uniformly balanced subset of data. This brief recalibration rectifies the logit disparities between old and new classes, ensuring balanced predictions across both old and newly introduced classes.

\begin{algorithm}[t]
\caption{KFN Incremental Learning Pipeline}
\label{alg:kfn}
\small
\begin{algorithmic}[1]
\Require Task sequence $\{T_1,\ldots,T_K\}$, threshold $\tau$, memory $\mathcal{M}$, loss weights $\lambda_{\text{CE}}, \lambda_{\text{rep}}, \lambda_{\text{KD}}, \lambda_{\text{feat}}$
\Ensure Global model $G$ retaining all prior task knowledge

\Statex \textbf{Phase 1 — Base Initialisation}
\State Train $G = \{\theta_{\text{global}}, \mathbf{W}_{\text{cls}}\}$ on $\mathcal{D}_1$ via $\mathcal{L}_{\text{CE}}$
\State $\mathcal{M} \leftarrow \text{sample}(\mathcal{D}_1)$

\Statex \textbf{Incremental Steps}
\For{$t = 2$ \textbf{to} $K$}

    \Statex \quad \textit{Phase 2 — Isolated Specialist Training}
    \State Initialise specialist $\phi_t$ with random weights
    \State Train $\phi_t$ on $\mathcal{D}_t$ with all $\phi_{<t}$, $\theta_{\text{global}}$ \textbf{frozen} \Comment{$\nabla_{\phi_{<t}}\mathcal{L} = 0$ strictly}

    \Statex \quad \textit{Novelty-Gated Capacity Decision}
    \State Draw reference set $\mathcal{R}_t \subset \mathcal{D}_t^{\text{train}}$, $|\mathcal{R}_t| = n_{\text{ref}}$
    \State $z_{\text{new}} \leftarrow \phi_t(x),\quad z_{\text{old}} \leftarrow G(\phi_{<t};\,\theta_{\text{global}}) \quad \forall x \sim \mathcal{R}_t$
    \State Pool and $\ell_2$-normalise descriptors; form basis $\hat{F}_{\text{old}}$; $P \leftarrow \hat{F}_{\text{old}}\hat{F}_{\text{old}}^{+}$ \Comment{Eqs.~(5)--(7)}
    \State $\bar{N} \leftarrow \mathbb{E}_{x \sim \mathcal{R}_t} \bigl[\|(I-P)\,\hat{f}_{\text{new}}(x)\|_2\bigr]$ \Comment{Bounded residual, Eqs.~(5)--(8)}
    \If{$\bar{N} > \tau$}
        \State $d_{\text{new}} \leftarrow d_{\text{high}}$;\quad $\alpha \leftarrow \max(0.1,\, 1-\bar{N})$ \Comment{Expand: allocate new channels; scale reuse}
    \Else
        \State $d_{\text{new}} \leftarrow 0$;\quad $\alpha \leftarrow \max(0.1,\, 1-\bar{N})$ \Comment{Reuse: no additional capacity}
    \EndIf

    \Statex \quad \textit{Phase 3 — Structural Fusion}
    \State \textbf{Freeze} $\phi_{<t}$;\quad \textbf{Unfreeze} $\phi_t$, $\theta_{\text{SFM}}$, $\mathbf{W}_{\text{cls}}$
    \State \textit{SFM forward pass:}
    \State \quad $\Delta z_{\text{old}} \leftarrow F_{\text{reuse}}(\phi_t(x);\,\theta_{\text{reuse}})$
    \State \quad $z_{\text{new}} \leftarrow F_{\text{expand}}(\phi_t(x);\,\theta_{\text{expand}})$
    \State \quad $z_t \leftarrow \bigl(z_{\text{old}} + \alpha\,\sigma(g)\cdot\Delta z_{\text{old}}\bigr) \oplus z_{\text{new}}$ \Comment{$\alpha$ from capacity decision; $\sigma(g)$ learned}
    \State Minimise $\mathcal{L} = \lambda_{\text{CE}}\mathcal{L}_{\text{CE}} + \lambda_{\text{rep}}\mathcal{L}_{\text{rep}} + \lambda_{\text{KD}}\mathcal{L}_{\text{KD}} + \lambda_{\text{feat}}\mathcal{L}_{\text{feat}}$ \Comment{$\nabla_{\phi_{<t}}\mathcal{L}_{\text{fusion}}=0$}

    \Statex \quad \textit{Phase 4 — Classifier Calibration}
    \State \textbf{Freeze} all feature and fusion parameters
    \State Fine-tune $\mathbf{W}_{\text{cls}}$ on balanced $\mathcal{M} \cup \text{sample}(\mathcal{D}_t)$ \Comment{Equal exemplars per class}

    \State $\mathcal{M} \leftarrow \mathcal{M} \cup \text{sample}(\mathcal{D}_t)$

\EndFor
\end{algorithmic}
\end{algorithm}

\subsection{Objective Function}
\label{subsec: Objective Function}
During the fusion phase, the global model is optimized to integrate novel task knowledge while preserving historical representations. This process is governed by a composite objective function, formulated as a weighted sum of four distinct constraints:
\begin{equation}
\label{eq:fusion_loss}
\mathcal{L} = \lambda_{CE}\mathcal{L}_{CE} + \lambda_{rep}\mathcal{L}_{rep} + \lambda_{KD}\mathcal{L}_{KD} + \lambda_{feat}\mathcal{L}_{feat}
\end{equation}

Each term enforces a specific functional or structural requirement, with the scalar coefficients explicitly controlling the trade-offs between plasticity and stability during optimization. Empirically, these coefficients are strictly bounded to prioritize new task acquisition while maintaining regularization (e.g., $\lambda_{CE}=2.5, \lambda_{rep}=0.8, \lambda_{KD}=1.0, \lambda_{feat}=1.2$).

The primary objective for the incoming task $T_t$ is formulated as the standard cross-entropy loss evaluated over the current dataset $\mathcal{D}_t$:
\begin{equation}
\mathcal{L}_{CE} = - \mathbb{E}_{(x,y)\sim \mathcal{D}_t} \left[ \sum_{c=1}^{C} y_c \log p_c(x) \right]
\end{equation}

This term actively drives discriminative learning, enabling the architecture to acquire novel task-specific decision boundaries.

To stabilize the mapping of previously learned tasks, we apply a replay loss computed over a bounded episodic memory $\mathcal{M}$ containing historical exemplars:
\begin{equation}
\mathcal{L}_{rep} = - \mathbb{E}_{(x,y)\sim \mathcal{M}} \left[ \sum_{c=1}^{C} y_c \log p_c(x) \right]
\end{equation}

By approximating the past data distribution, this objective reduces logit drift and heavily penalizes deviation of the classifier head on prior domains.

Further functional stability is enforced via knowledge distillation, which aligns the predictive distribution of the actively learning model with that of the frozen global model from step $t-1$. To capture inter-class structural similarities, this formulation incorporates temperature scaling:
\begin{multline}
\label{eq:kd_loss}
\mathcal{L}_{\text{KD}} = \mathbb{E}_{x \sim \mathcal{D}_t \cup \mathcal{M}} \Bigg[ T^2 \cdot \text{KL}\Bigg( \operatorname{softmax}\left(\frac{s_{\text{old}}(x)}{T}\right) \\
\Big\Vert \operatorname{softmax}\left(\frac{s_{\text{new}}(x)}{T}\right) \Bigg) \Bigg]
\end{multline}

where $\mathrm{KL}$ denotes the Kullback--Leibler divergence and $T$ is a temperature hyperparameter (e.g., $T=2.0$). This constraint rigorously preserves the functional output mapping established by preceding tasks. The multiplier $T^2$ is the sole KL weighting factor: with $T = 2.0$ this equals $4.0$, which is the ``KL weighting factor of 4.0'' referenced in Section~\ref{subsec: Implementation Details and Reproducibility}---the two quantities are the same factor, not independent coefficients. In the implementation, the expectation in Eq.~\eqref{eq:kd_loss} is realized over a mixed batch drawn from the current task data $\mathcal{D}_t$ interleaved with exemplars from the replay buffer $\mathcal{M}$: the current-task component anchors plasticity while the replayed component anchors the teacher--student alignment on prior-task inputs, so both distributions contribute to the distilled expectation.

Crucially, to maintain topological integrity within the latent space, KFN employs a feature consistency penalty. This objective minimizes the squared Euclidean distance between the current global memory representation $z_{\text{old}}^{\text{new}}$ and the historical target $z_{\text{old}}^{\text{frozen}}$ generated by the frozen teacher network:
\begin{equation}
\mathcal{L}_{feat} = \mathbb{E}_{x \sim \mathcal{M}} \left[ \left\| z_{\text{old}}^{\text{new}}(x) - z_{\text{old}}^{\text{frozen}}(x) \right\|_2^2 \right]
\end{equation}

This spatial regularizer explicitly enforces consistency in the global representation space. It ensures that the reusable component extracted by the Structural Fusion Module integrates smoothly, actively preventing structural drift in $z_{\text{old}}$ as new specialist features are actively fused.

These auxiliary objectives act as stabilization constraints; however, the primary resistance to catastrophic forgetting in KFN arises from its structural decoupling of representations, rather than from loss-based regularization.

The episodic memory buffer $\mathcal{M}$ used in the replay and distillation terms is populated via uniform random sampling of historical task data, strictly bounded to $M = 20$ exemplars per class to simulate tight memory constraints typical of realistic continual learning scenarios. The buffer is refreshed upon the introduction of each new task: the existing exemplars are retained, and new exemplars from the current task are sampled and appended up to the per-class budget. Through the combination of this memory-anchored regularization and the physical decoupling of the network pathways, KFN empirically maintains the loss on prior tasks within controlled tolerance bounds throughout sequential deployment. We emphasize that the exact bound $\Delta\mathcal{L}_{<t} \leq \epsilon$ depends on the interplay between buffer size, distillation strength, and task similarity, and we treat it as an empirical property validated by the retention metrics rather than a formal guarantee.

\subsection{Complexity Analysis}
\label{subsec: Complexity Analysis}
A fundamental limitation of modular and expandable architectures is the asymptotic scaling of memory and computational overhead. We give an exact accounting for KFN, separating shared parameters from genuinely per-task modules.

\textbf{Memory Complexity:} Because every specialist is retained and participates in inference, a lower bound on the deployed parameter count is
\begin{equation}
P_T \;\geq\; P_1 + (T-1)\,|\phi| \;=\; \Omega(T),
\label{eq:linear_growth}
\end{equation}
where $|\phi|$ denotes the fixed parameter footprint of a single specialist network. The novelty gate can reduce only the expansion-branch increment $d_{\text{new}}^{(t)}$ to zero; it cannot remove the per-task specialist increment, so growth is at least linear for any task sequence under the evaluated architecture, and strictly linear in the benchmark sequences evaluated here where every task exceeds $\tau$. A complete decomposition separates the shared state from per-task modules:
\begin{equation}
P_{\text{total}}(T) = P_{\text{shared}}(T) + P_{\text{cls}}(C_T) + \sum_{t=2}^{T}\bigl(P_{\text{spec},t} + P_{\text{proj},t} + P_{\text{expand},t}\bigr),
\label{eq:param_decomp}
\end{equation}
where $P_{\text{shared}}(T)$ comprises the gate, bottleneck, and historical specialist projections (with the gate and bottleneck reused rather than duplicated per task, and the projection list growing by one module per task), $P_{\text{cls}}(C_T)$ denotes the classifier growth with the number of observed classes $C_T$, $P_{\text{spec},t} = |\phi|$ is the retained specialist, $P_{\text{proj},t}$ is the specialist-specific projection, and $P_{\text{expand},t} = \sum_{\ell} (k_h k_w d_{\text{in}}^{(\ell)} d_{\text{out}}^{(\ell)} + d_{\text{out}}^{(\ell)})$ sums the convolutional parameters of the expansion branch (for an expansion layer with kernel size $k_h \times k_w$, input width $d_{\text{in}}$, and output width $d_{\text{out}}$; the quantity $d_{\text{new}} \cdot h \cdot w$ is an activation volume, not a parameter count). Unlike progressive architectures that unconditionally instantiate dense parallel backbones ($|\phi_{\mathrm{PNN}}| \approx 11.2\mathrm{M}$ per task), KFN's specialists are lightweight ($|\phi| \approx 2.8\mathrm{M}$), yielding a 4$\times$ per-task parameter advantage over naive expansion.

\textbf{Computational Complexity (Training):} Standard joint-training and deeply entangled continual learning methods require gradient computation through the entire global backbone, forcing the backward pass compute cost to scale linearly or quadratically with the number of tasks. KFN structurally partitions optimization. During independent specialist training (Phase 2), the backward pass strictly evaluates the parameters of the active specialist, $w_t$. Because $\frac{\partial \phi_{<t}}{\partial w_t} = \mathbf{0}$, gradients are never routed through historical specialists. Consequently, the Phase-2 optimization compute cost per task remains $\mathcal{O}(1)$ with respect to prior tasks; the full incremental pipeline additionally includes Phase-3 forward evaluation and aggregation of all historical specialists, so the per-task pipeline cost inherits the $\Omega(T)$ term of Eq.~\eqref{eq:linear_growth} at late stages of the sequence.

\textbf{Computational Complexity (Inference \& Parallelizability):} During the forward pass, the model must evaluate all $T$ retained specialists, so the total inference arithmetic work satisfies
\begin{equation}
W_{\text{infer}}(T) \;\geq\; \sum_{t=1}^{T} F_{\text{spec},t} \;=\; \Omega(T),
\label{eq:infer_work}
\end{equation}
where $F_{\text{spec},t}$ denotes the FLOPs of specialist $t$. Because the parameter sets $w_1, \dots, w_T$ are disjoint and functionally independent until the global fusion bottleneck, the feature extraction pathways share no inter-layer dependencies and are perfectly parallelizable across specialist modules. On hardware with sufficient accelerator capacity, memory bandwidth, and scheduling resources, the effective forward-pass \emph{wall-clock latency} can approach $\mathcal{O}(\text{FLOPs}(\phi))$; this is a resource-dependent property, not an asymptotic guarantee---constant latency under growing $T$ presumes that the deployed accelerator provision scales accordingly. Total arithmetic work remains $\Omega(T)$ under any fixed resource budget, and this is accounted for in the wall-clock measurements reported in Section~\ref{subsec: Computational efficiency and linear parameter scaling}.

\subsection{Implementation Details and Reproducibility}
\label{subsec: Implementation Details and Reproducibility}
All models were implemented in PyTorch and trained on NVIDIA GPUs with mixed-precision training (AMP). For the ten-task Split CIFAR-100 protocol, the base model (Phase~1) was trained for 10 epochs using the AdamW optimizer with a learning rate of $1 \times 10^{-3}$. Subsequent specialists (Phase~2) were trained for 30 epochs with a learning rate of $1 \times 10^{-3}$ and weight decay of $1 \times 10^{-4}$. The structural fusion phase (Phase~3) utilized a differential learning rate strategy over 40 epochs: newly added fusion and classifier components were optimized at $3 \times 10^{-3}$, while historical gating and bottleneck components were updated at a reduced rate of $5 \times 10^{-4}$, both with weight decay of $1 \times 10^{-4}$. Phased gradient unfreezing was applied during Phase~3: the old bottleneck was released at epoch 0, and the old gate was released at epoch 20. The bias correction step (Phase~4) was executed over 5 epochs with a learning rate of $5 \times 10^{-5}$.

\textbf{Task-boundary assumption.} KFN instantiates and trains a new specialist upon the arrival of each task; training therefore assumes \emph{known task boundaries}. This is the standard training-time supervision setting used by the architecture-growing methods KFN is compared with (PNN, DEN). Task identity is \emph{not} used at inference: all Class-IL evaluations select the argmax over all currently exposed class logits without task masking. Boundary-detection or task-free variants are left to future work.

The fusion loss coefficients (cf.\ Eq.~\ref{eq:fusion_loss}) were set as $\lambda_{\mathrm{CE}} = 2.5$, $\lambda_{\mathrm{rep}} = 0.8$, $\lambda_{\mathrm{KD}} = 1.0$, and $\lambda_{\mathrm{feat}} = 1.2$. Knowledge distillation employed a temperature of $T = 2.0$; the resulting KL divergence weighting factor of $T^2 = 4.0$ is this temperature scaling (Section~\ref{subsec: Objective Function}), not an independent coefficient. The cosine classifier's learnable temperature parameter $\sigma$ was initialized to 10.0. Data augmentation for all methods included random cropping with 4-pixel padding, random horizontal flipping, and color jittering (brightness, contrast, and saturation each perturbed by $\pm 0.2$).

To ensure fair comparison, all shared-parameter baselines (LwF, Replay, LwF$^{++}$, Replay$^{++}$) were implemented with equivalent ResNet-based backbone architectures ($\approx$11.2M parameters) and bounded to the same episodic memory budget of 20 exemplars per class where applicable. Vanilla variants used standard cross-entropy or distillation losses without augmentation; the $+$ tier added a replay buffer; the $++$ tier additionally incorporated supervised contrastive loss, Mixup augmentation, and an exponential moving average teacher (Replay$^{++}$), or dual-temperature distillation with amplified replay (LwF$^{++}$). All experiments (including EWC in Table~\ref{tab:2task}) were repeated across all 5 prespecified random seeds. Deterministic cuDNN settings were enabled; deterministic execution reproduces a fixed run exactly but does not eliminate variation across initialization seeds or class orderings, so between-method comparisons are reported as mean $\pm$ standard deviation with seed-level dispersion disclosed per table. The same seed set and class ordering was used across all methods within a given table to enable paired comparison. Where a table reports a single run, the caption discloses it explicitly and labels the result as exploratory (Table~\ref{tab:10task}).

\section{Experimental Results}
\label{sec:Results}
\subsection{KFN achieves balanced long-horizon class-incremental learning}
\label{subsec: KFN achieves balanced long-horizon class-incremental learning}
We evaluate KFN on two class-incremental learning protocols of increasing difficulty: a two-task split of CIFAR-10 (five classes per task) and a ten-task split of CIFAR-100 (ten classes per task). In both settings, tasks are presented sequentially without full access to prior task datasets; a bounded exemplar memory of $M = 20$ exemplars per class is retained for replay and distillation objectives, simulating tight memory constraints typical of realistic continual learning deployments. We report the average accuracy across all observed tasks after the final task, $\bar{\mathcal{A}} = \frac{1}{T}\sum_{t=1}^{T} \mathcal{A}_{T,t}$, where $\mathcal{A}_{i,j}$ denotes accuracy on task $j$ after learning task $i$. To characterize the stability--plasticity balance, we additionally report three complementary metrics: \emph{plasticity}, defined as the mean diagonal accuracy $\frac{1}{T}\sum_{t=1}^{T} \mathcal{A}_{t,t}$ (how well each task is learned upon first encounter); \emph{forgetting} $\mathcal{F} = \frac{1}{T-1}\sum_{t=1}^{T-1}\bigl(\max_{i \geq t} \mathcal{A}_{i,t} - \mathcal{A}_{T,t}\bigr)$ (the average accuracy loss on prior tasks); and the standard deviation of final per-task accuracies $\sigma(\mathcal{A})$ (the uniformity of performance across tasks). All KFN results are reported as mean $\pm$ standard deviation over 5 random seeds; where appropriate, we note that the performance differences between KFN and the strongest baselines exceed two standard deviations of either method's distribution. We emphasise that a two-standard-deviation comparison is a descriptive dispersion check, not a formal hypothesis test; deterministic cuDNN settings reproduce a fixed run but do not eliminate seed- and class-order-level variation, so we treat seed-level dispersion as the primary uncertainty indicator throughout and avoid inferential claims beyond it. We compare KFN against four baselines spanning the major families of continual learning methods: Experience Replay (ER), Learning without Forgetting (LwF), Elastic Weight Consolidation (EWC), and Mixture of Experts (MoE).

\begin{figure*}[h]
\centering
\begin{minipage}{0.48\textwidth}
  \centering
  \includegraphics[width=\linewidth]{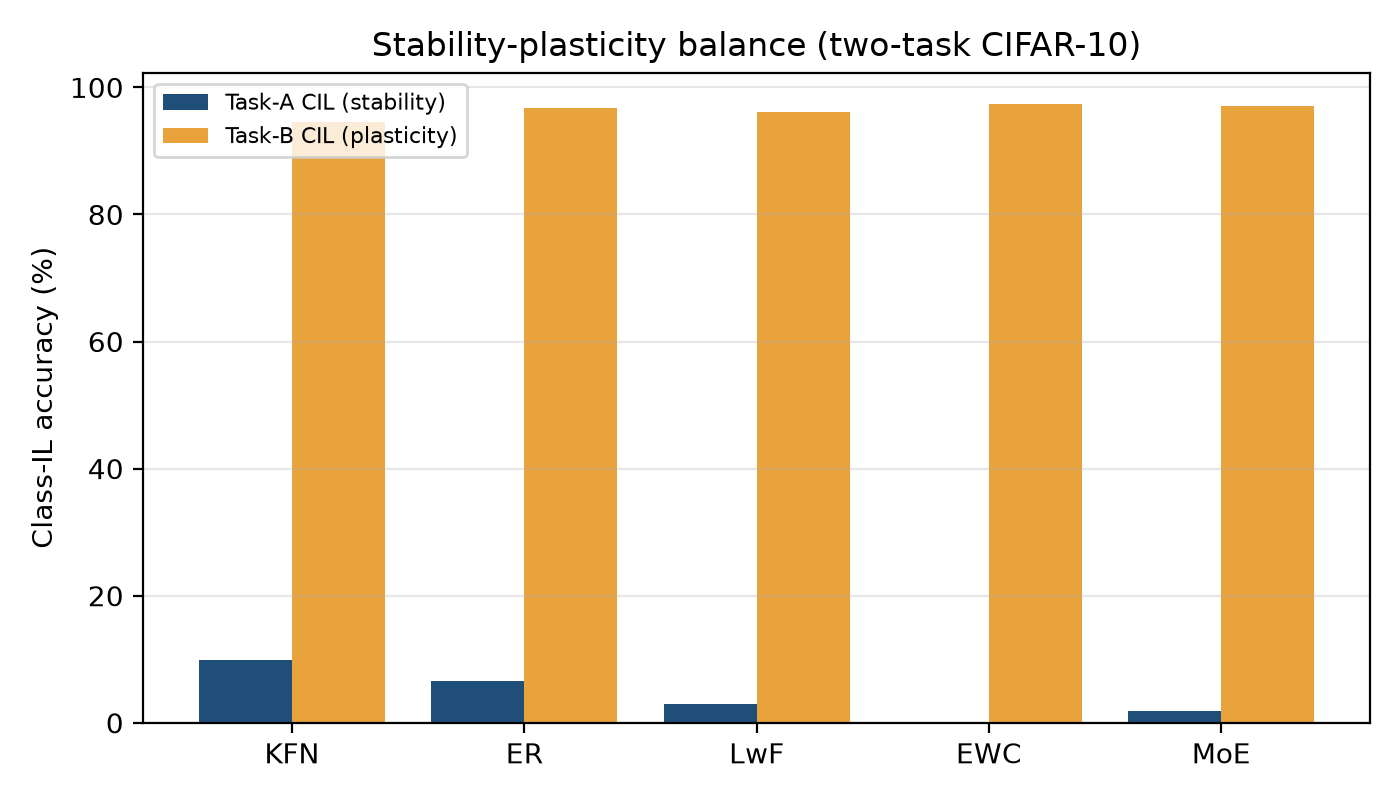}
  \caption{\textbf{Stability-plasticity tradeoff.} Performance on the two-task CIFAR-10 protocol, illustrating KFN's ability to maximize plasticity without sacrificing historical stability.}
  \label{fig:stability_plasticity_tradeoff}
\end{minipage}\hfill
\begin{minipage}{0.48\textwidth}
  \centering
  \includegraphics[width=\linewidth]{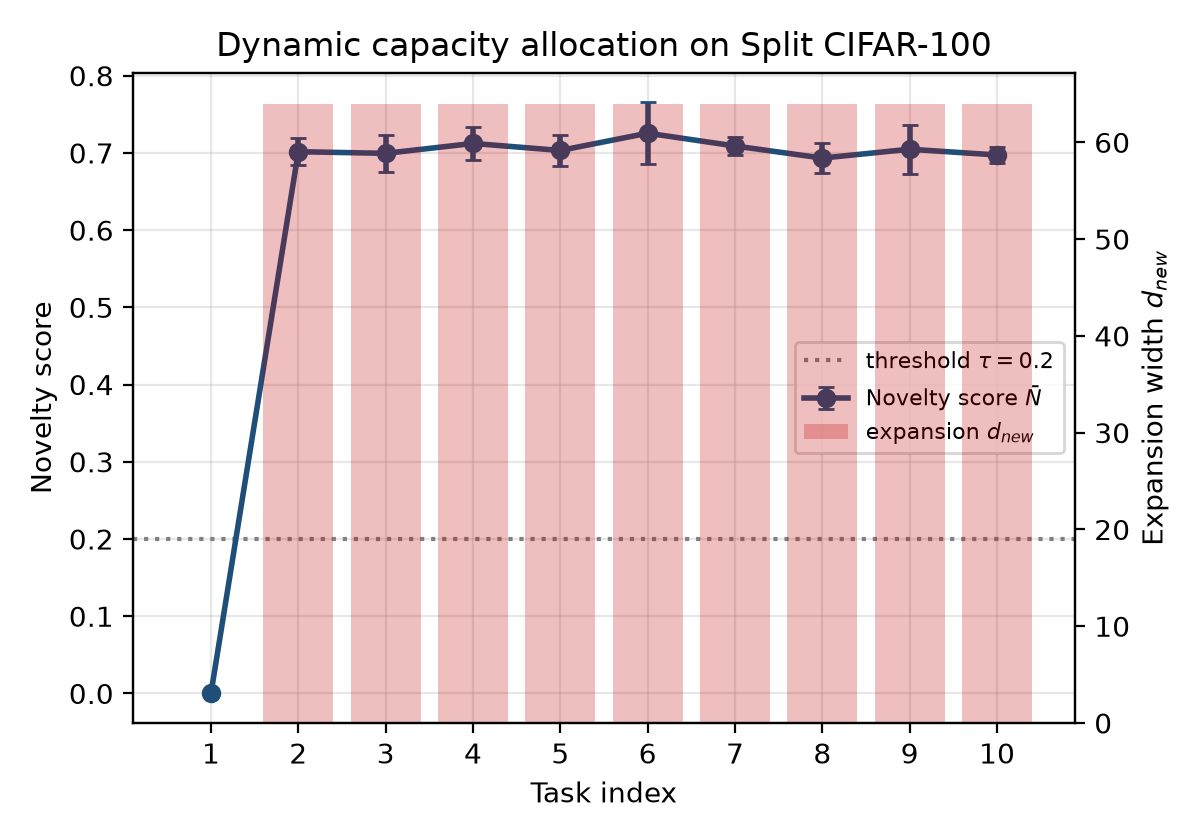}
  \caption{\textbf{Dynamic capacity allocation.} The novelty score trajectory and the resulting uniform structural expansion across the ten-task CIFAR-100 sequence.}
  \label{fig:novelty_trajectory}
\end{minipage}
\end{figure*}

\begin{table}[h]
\centering
\caption{\textbf{Two-task class-incremental learning on CIFAR-10.} Each task comprises five classes. KFN is the only method that simultaneously achieves high accuracy on both the retained task and the newly learned task. ALL RESULTS ARE REPORTED AS MEAN $\pm$ STANDARD DEVIATION OVER 5 RANDOM SEEDS. ALL 5 prespecified EWC seeds are reported. Forgetting is measured as $\mathcal{A}_A^{\mathrm{init}} - \mathcal{A}_A^{\mathrm{final}}$. Bold: best; \underline{underline}: second best.}
\label{tab:2task}
\resizebox{\columnwidth}{!}{%
\begin{tabular}{lccccc}
\toprule
Method & $\mathcal{A}_A^{\mathrm{init}}$ & $\mathcal{A}_A^{\mathrm{final}}$ & $\mathcal{A}_B^{\mathrm{final}}$ & $\bar{\mathcal{A}}$ & Forgetting \\
\midrule
KFN (ours) & $\mathbf{75.76 \pm 1.80}$ & $\mathbf{9.96 \pm 1.53}$ & $\mathbf{94.50 \pm 0.26}$ & $\mathbf{52.23}$ & $\mathbf{65.80 \pm 2.05}$ \\
\midrule
ER   & $69.31 \pm 1.64$ & $6.60 \pm 0.85$ & $96.81 \pm 0.35$ & $51.70$ & $62.71 \pm 1.62$ \\
LwF  & $\mathbf{94.02 \pm 0.17}$ & $3.00 \pm 0.36$ & $96.19 \pm 0.11$ & $49.60$ & $91.02 \pm 0.48$ \\
EWC  & $40.94 \pm 9.60$ & $0.00 \pm 0.00$ & $\mathbf{97.44 \pm 0.15}$ & $48.72$ & $40.94 \pm 9.60$ \\
MoE  & $63.30 \pm 2.46$ & $1.93 \pm 2.63$ & $97.01 \pm 0.16$ & $49.47$ & $61.37 \pm 1.92$ \\
\bottomrule
\end{tabular}%
}
\end{table}

\paragraph{Two-task evaluation reveals a universal stability--plasticity imbalance.}
The two-task CIFAR-10 results (Table~\ref{tab:2task}) establish a clear diagnostic of how each method family behaves under strict $m=20$ Class-IL evaluation. Because the output space expands without task-identity masking at test time, all shared-parameter baselines suffer severe recency bias toward Task~B (96.19\%--97.44\%) while collapsing on Task~A. EWC catastrophically forgets Task~A (0.00\% across all 5 seeds) despite achieving the highest Task~B accuracy (97.44\%), illustrating the failure mode of parameter-level regularization when new classifier heads dominate the unified decision boundary. LwF and MoE similarly collapse on Task~A under unmasked Class-IL evaluation (retaining only 3.00\% and 1.93\%, respectively), while ER retains 6.60\% on Task~A. KFN achieves the highest Class-IL accuracy on the retained task (9.96\%---51\% higher than ER and more than 3$\times$ LwF) and the highest overall average accuracy (52.23\%), while simultaneously achieving 94.50\% on the new task (Fig.~\ref{fig:stability_plasticity_tradeoff}).

\begin{table}[h]
\centering
\caption{\textbf{Ten-task class-incremental learning on Split CIFAR-100.} Each task comprises ten classes presented sequentially. $\bar{\mathcal{A}}$: average final accuracy across all ten tasks. Plasticity: mean accuracy on each task immediately after learning it (diagonal of the accuracy matrix). $\mathcal{F}$: average forgetting. $\sigma(\mathcal{A})$: standard deviation of final per-task accuracies, measuring performance uniformity. Min~$\mathcal{A}$: worst single-task accuracy after all ten tasks. All methods use identical backbone architectures (11.2M parameters). \emph{Exploratory single-run disclosure:} due to the computational cost of ten sequential tasks per method, each row reports one run per method under an identical seed and class ordering; these values are not seed-averaged and should be read as exploratory rather than as mean~$\pm$~std estimates (cf.\ Section~\ref{subsec: Implementation Details and Reproducibility}). EWC and MoE are omitted from the ten-task evaluation; given their failure modes on the two-task protocol (complete forgetting and severe plasticity deficit, respectively), their degradation at scale is predictable. Bold: best; \underline{underline}: second best.}
\label{tab:10task}
\begin{tabular}{lccccc}
\toprule
Method & $\bar{\mathcal{A}}$ $\uparrow$ & Plasticity $\uparrow$ & $\mathcal{F}$ $\downarrow$ & $\sigma(\mathcal{A})$ $\downarrow$ & Min~$\mathcal{A}$ $\uparrow$ \\
\midrule
KFN (ours) & \textbf{22.29} & 74.41 & $\mathbf{57.92}$ & $\mathbf{16.73}$ & $\mathbf{11.94}$ \\
LwF        & 9.07 & $\underline{83.89}$ & 83.14 & 26.56 & 0.00 \\
Replay++   & 8.91 & $\mathbf{85.48}$ & 85.08 & 26.73 & 0.00 \\
Replay     & $\underline{16.21}$ & 82.31 & 73.44 & $\underline{21.66}$ & $\underline{2.44}$ \\
\bottomrule
\end{tabular}
\end{table}

\paragraph{Long-horizon evaluation confirms KFN's balanced performance at scale.}
We validate these findings on the substantially more challenging ten-task Split CIFAR-100 benchmark, where the model must sequentially learn 100 classes organized into ten disjoint tasks of ten classes each. The results are summarized in Table~\ref{tab:10task}. KFN achieves an average final accuracy of 22.29\%, above Replay (16.21\%) and substantially above LwF (9.07\%) and Replay++ (8.91\%). However, the aggregate accuracy alone understates a critical qualitative difference in how this performance is distributed across tasks.

KFN is the only method that maintains both high plasticity and bounded forgetting. Its mean diagonal accuracy (74.41\%) is comparable to the baselines' (82--85\%), but the baselines pay for this plasticity with catastrophic forgetting (73--85\%), while KFN's forgetting is 57.92\%---15.5 pp lower than the next best. After all ten tasks, KFN's per-task accuracies yield a standard deviation of 16.73~pp---lower than LwF ($\sigma = 26.56$) and Replay++ ($\sigma = 26.73$). The minimum single-task accuracy after all ten tasks is 11.94\% for KFN, compared to 0.00\% for LwF and Replay++ and 2.44\% for Replay. In other words, KFN's worst-performing task remains well above chance and above every baseline's worst task, whereas every baseline leaves at least one task in a near-degenerate state.

\paragraph{Distillation without structural isolation suffers severe recency bias at scale.}
Under strict Class-IL evaluation, LwF exhibits high plasticity (83.89\%) on incoming tasks but suffers catastrophic forgetting (83.14\%) across prior tasks, leaving its worst task at 0.00\%. LwF's Task~10 accuracy reaches 88.74\%---the newest task is well learned---but at the cost of 83.14\% forgetting on all prior tasks. KFN, by contrast, achieves 71.16\% on Task~10 with a forgetting rate of 57.92\%, 15.5 pp lower than the next best. The apparent stability of old tasks under distillation-based methods comes at the cost of systematically failing to preserve them.

Vanilla Replay exhibits a complementary failure: it achieves high plasticity (82.31\%) but forgets 73.44\% of prior-task knowledge, yielding an average accuracy of only 16.21\% and leaving at least one task at 2.44\%. Replay++, which augments the replay buffer with distillation-based regularization, actually performs worse (8.91\% average accuracy) with the highest forgetting (85.08\%) and a minimum task accuracy of 0.00\%, demonstrating that adding more loss terms within a fixed-capacity architecture does not resolve the underlying capacity saturation.

\begin{figure}[h]
\centering
\includegraphics[width=\linewidth]{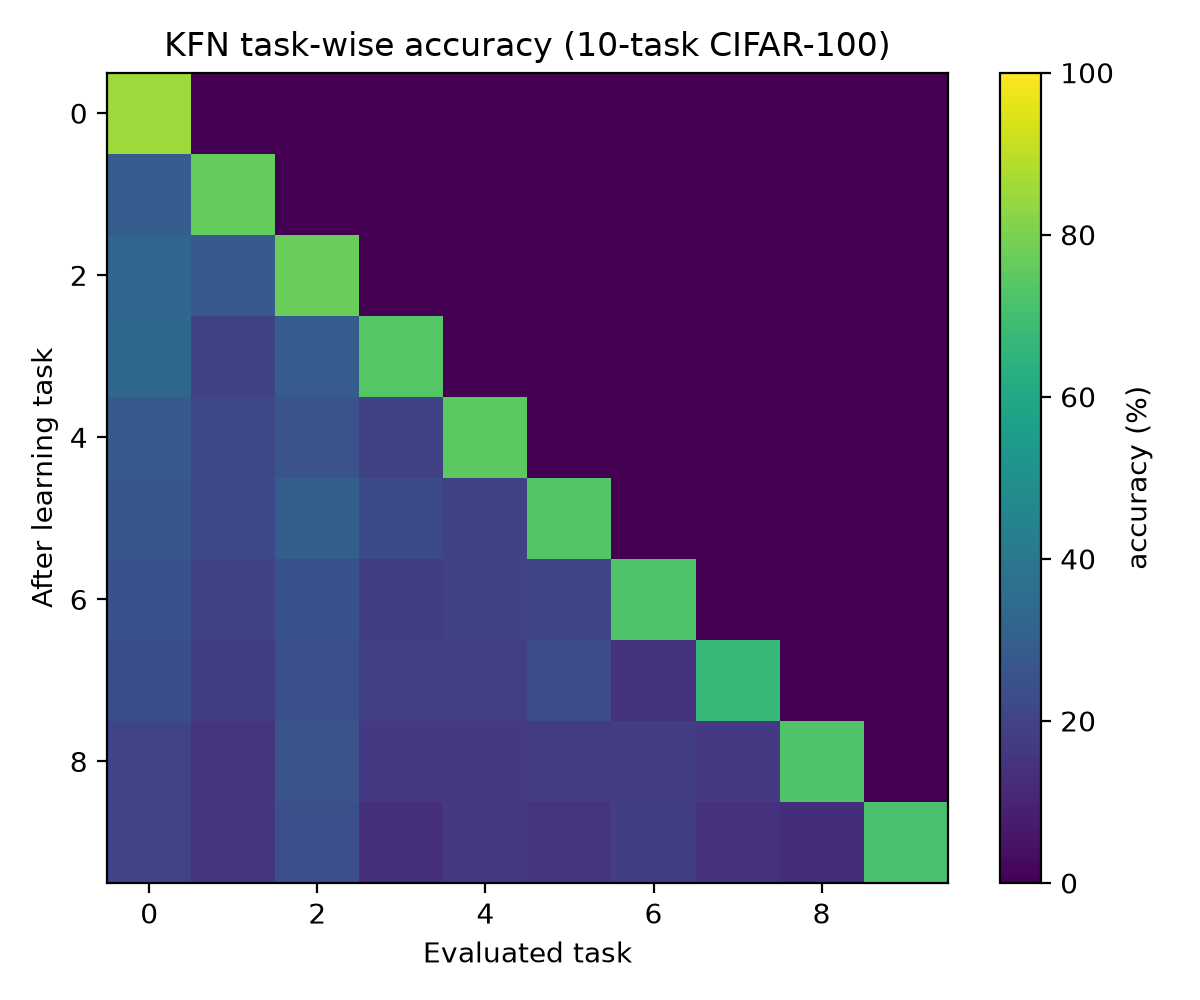}
\caption{\textbf{Task-wise accuracy progression.} Heatmap illustrating the evolution of per-task accuracy across the 10-task sequential learning process. KFN demonstrates a controlled, graded forgetting profile compared to the catastrophic interference observed in baselines.}
\label{fig:accuracy_heatmap}
\end{figure}

\paragraph{KFN's novelty-driven expansion allocates capacity proportional to task dissimilarity.}
The ten-task evaluation reveals an important property of KFN's integration mechanism. Figure~\ref{fig:novelty_trajectory} plots the novelty score computed at each incremental step alongside the resulting structural expansion. Upon encountering Task~2, the novelty score is 0.702, triggering a 64-channel expansion. As the task sequence progresses, novelty scores remain consistently in the range $[0.65, 0.73]$, with minor local variation (Task 9: 0.705, Task 8: 0.693) reflecting stochastic variation in how CIFAR-100 super-classes partition the feature space; the realised SVD rank of the reference basis is 32 at every step, confirming a stable, well-conditioned projector. Importantly, the expansion dimension remains constant at 64 channels across all tasks, suggesting that the representational requirements per task are approximately uniform even as the novelty of incoming tasks increases. This controlled expansion ensures that KFN allocates new capacity only when structurally necessary---preventing both the representational rigidity of fixed-capacity methods and the unbounded growth of naive architectural approaches.

The resulting forgetting trajectory under KFN exhibits a graded pattern rather than the binary retention/collapse seen in baselines (Fig.~\ref{fig:accuracy_heatmap}): the standard deviation of final per-task accuracies is 16.73~pp, and the minimum remains at 11.94\%, confirming that no task is driven to complete collapse. This is qualitatively distinct from LwF and Replay++, whose minimum task accuracies reach 0.00\%. Critically, even the most forgotten task under KFN (11.94\% minimum) remains well above chance and substantially above the worst-performing task under any baseline.

\subsection{KFN outperforms shared-parameter continual learning baselines}
\label{subsec: KFN outperforms shared-parameter continual learning baselines}

The results in subsection \ref{subsec: KFN achieves balanced long-horizon class-incremental learning} compared KFN against standard continual learning baselines. However, a critical question remains: can shared-parameter methods close the gap with KFN when equipped with modern representation learning techniques? To answer this, we evaluate baselines at three tiers of augmentation: \emph{vanilla} (the original algorithms as proposed, e.g., pure LwF with standard knowledge distillation and no replay memory; pure Replay using only cross-entropy loss); \emph{intermediate} ($+$, standard modernizations such as basic data augmentation or a simple memory buffer added to a method that did not originally include one); and \emph{heavily aided} ($++$, methods pushed to their limits using state-of-the-art regularizers---specifically, LwF$++$ augments the distillation objective with a replay buffer and dual-temperature knowledge distillation, effectively converging toward Dark Experience Replay~\cite{Buzzega2020DER}, while Replay$++$ adds supervised contrastive loss (SupCon), Mixup augmentation, and an exponential moving average (EMA) teacher). This tiered design ensures that KFN is evaluated against the strongest possible shared-parameter competitors, not merely canonical implementations from prior work. All baselines at every tier share the same backbone architecture (11.2M parameters) and differ only in their training objectives and data access.

\paragraph{Class-incremental evaluation exposes a fundamental limitation invisible to task-aware metrics.}
We evaluate all tiers on a two-task CIFAR-100 protocol (50 classes per task) under both task-aware (TA) and class-incremental (CIL) evaluation. Under TA evaluation---where the model is informed of the task identity and predicts only among the classes of that task---vanilla LwF appears competent (74.20\% / 68.80\% on Tasks~A/B). Under CIL evaluation---where the model must discriminate among all 100 classes without task identity---the same model collapses on Task~A (7.49\%), retaining only 10.1\% of its initial TA accuracy. This 66.71~pp gap between TA and CIL accuracy on the old task reveals that vanilla LwF's apparent stability is an artifact of the evaluation protocol: the distillation loss preserves the relative ordering of old-class logits, but the absolute logit scale shifts dramatically in favor of new classes, causing systematic misclassification under the unified decision boundary required by CIL.

\begin{figure}[h]
\centering
\includegraphics[width=\linewidth]{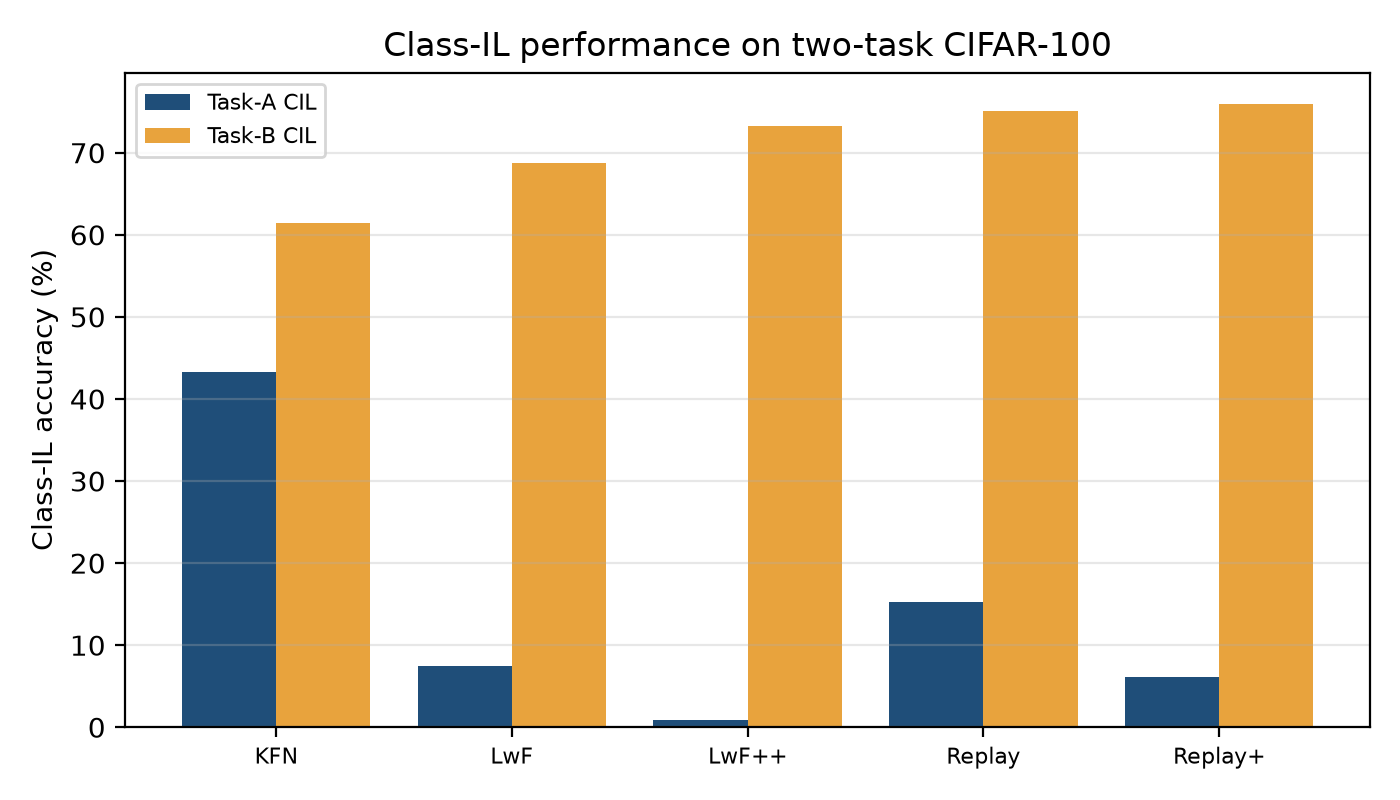}
\caption{\textbf{Effect of modern augmentations on CIL performance.} The bar chart illustrates the asymmetric impact of advanced regularizers on shared-parameter methods. While LwF$++$ achieves high Task~B accuracy at the cost of Task~A collapse, and Replay$+$ boosts Task~B by collapsing Task~A, KFN achieves optimal, symmetric performance across both domains without requiring loss-level augmentation.}
\label{fig:tier_progression}
\end{figure}

\begin{table*}[t] 
\centering
\caption{\textbf{Two-task CIFAR-100 under task-aware and class-incremental evaluation.} The tier column indicates the augmentation level (see text). CIL Retention measures the fraction of initial task-aware accuracy preserved under CIL evaluation: $\mathcal{A}_A^{\mathrm{CIL}} / \mathcal{A}_A^{\mathrm{init,TA}} \times 100$. ALL RESULTS ARE REPORTED AS MEAN $\pm$ STANDARD DEVIATION OVER 5 RANDOM SEEDS. CIL Retention is computed as the ratio of means, $\mathcal{A}_A^{\mathrm{CIL}} / \mathcal{A}_A^{\mathrm{init,TA}} \times 100$, applied to the displayed mean accuracies. Bold: best; \underline{underline}: second best.}
\label{tab:cifar100_2task_cil}
\begin{tabular}{llcccccc}
\toprule
 & & \multicolumn{2}{c}{Task-aware} & \multicolumn{2}{c}{Class-incremental} & \\
\cmidrule(lr){3-4} \cmidrule(lr){5-6}
Method & Tier & $\mathcal{A}_A$ & $\mathcal{A}_B$ & $\mathcal{A}_A$ & $\mathcal{A}_B$ & Avg CIL & CIL Ret. \\
\midrule
\textbf{KFN (ours)} & --- & $\mathbf{59.39 \pm 1.18}$ & $67.57 \pm 0.72$ & $\mathbf{43.24 \pm 3.90}$ & $61.43 \pm 2.76$ & $52.34$ & $\mathbf{72.7\%}$ \\
\midrule
LwF       & Vanilla       & $\mathbf{74.20 \pm 0.42}$ & $68.80 \pm 0.28$ & $7.49 \pm 0.50$ & $68.78 \pm 0.28$ & $38.14$ & $10.1\%$ \\
LwF$++$   & Heavily aided & $71.34 \pm 0.19$ & $73.21 \pm 0.73$ & $0.94 \pm 0.77$ & $73.21 \pm 0.72$ & $37.07$ & $1.3\%$ \\
\midrule
Replay    & Vanilla       & $51.54 \pm 0.31$ & $75.69 \pm 0.56$ & $\underline{15.24 \pm 0.96}$ & $75.10 \pm 0.50$ & $\mathbf{45.17}$ & $\underline{29.6\%}$ \\
Replay$+$ & Intermediate  & $44.05 \pm 1.40$ & $\mathbf{76.34 \pm 0.70}$ & $6.06 \pm 1.19$ & $\mathbf{75.96 \pm 0.55}$ & $41.01$ & $13.8\%$ \\
\bottomrule
\end{tabular}
\end{table*}

The results are summarized in Table~\ref{tab:cifar100_2task_cil}. KFN achieves the highest CIL retention (72.7\%), retaining 43.24\% of its initial Task-A accuracy under the unified 100-class decision boundary, whereas the best baseline (Replay) retains only 29.6\% and LwF$++$ collapses to 1.3\%. KFN's TA-to-CIL gap is 16.15~pp on Task~A and 6.14~pp on Task~B. This relative stability of old-task representations under a growing decision boundary is the primary structural effect of dedicated channel allocation. LwF$++$ achieves a CIL retention of only 1.3\%--- that the combination of distillation, replay, and dual-temperature amplification within a fixed-capacity architecture drives old-class representations to complete collapse under CIL evaluation, despite achieving the highest Task-B accuracy (73.21\%). Replay$+$ exhibits the most extreme asymmetry: its Task~A CIL accuracy (6.06\%) is 38.0~pp below its TA accuracy (44.05\%), while its Task~B gap is negligible (0.38~pp), confirming that the contrastive augmentation improves new-task representation at the direct expense of old-task discriminability under CIL. All CIL Retention values in Tables~\ref{tab:cifar100_2task_cil} and \ref{tab:eurosat} are computed as the ratio of displayed means, $\mathcal{A}_A^{\mathrm{CIL}} / \mathcal{A}_A^{\mathrm{init,TA}} \times 100$; we adopt the ratio-of-means aggregation and state it explicitly here to keep the derived percentages reproducible from the table entries.

\paragraph{Modern enhancements yield inconsistent and method-dependent gains.}
The transition from vanilla to enhanced variants produces strikingly asymmetric outcomes across method families. LwF$++$, which augments distillation with a replay buffer and amplified dual-temperature knowledge distillation, drives Task~A CIL accuracy from 7.49\% to 0.94\%---a further loss of 6.55~pp---and reduces CIL retention from 10.1\% to 1.3\%. This demonstrates that additional loss terms within a fixed-capacity architecture do not resolve the underlying representational competition; instead, the added optimisation pressure further displaces old-class representations. Replay$+$, however, exhibits the opposite pattern: its Task~A CIL accuracy \emph{decreases} from 15.24\% to 6.06\% ($-$9.18~pp), and CIL retention drops from 29.6\% to 13.8\%, despite substantial gains in Task~B accuracy (75.10\% $\to$ 75.96\%). The supervised contrastive loss and Mixup augmentation that define the $+$ tier improve new-task representation learning but simultaneously disrupt the logit calibration of old-class outputs, widening the inter-task logit gap that drives CIL misclassification \cite{khosla2020supcon,zhang2018mixup}. 

These divergent outcomes reveal that the gains from modern regularizers are not monotonic: they depend on whether the augmentation directly addresses the CIL failure mode (as LwF$++$ does via replay-anchored distillation) or inadvertently exacerbates logit imbalance by over-optimizing new task features (as Replay$+$ does via contrastive augmentation). Ultimately, neither augmentation strategy successfully resolves the stability-plasticity dilemma across both tasks simultaneously, confirming that the performance ceiling is architecturally bound rather than merely a function of loss engineering.

\subsection{Structural isolation mechanisms preserve prior task knowledge}
\label{subsec: Structural isolation mechanisms preserve prior task knowledge}

The results in \ref{subsec: KFN achieves balanced long-horizon class-incremental learning} and \ref{subsec: KFN outperforms shared-parameter continual learning baselines} demonstrate that KFN achieves balanced continual learning where shared-parameter methods fail. We now investigate the mechanistic basis of this advantage through a controlled ablation study. KFN's architecture implements three complementary forms of isolation: \emph{channel-space isolation}, which dedicates separate channel groups to old and new task representations (isolation in the channel dimension, not in the $h \times w$ spatial grid); \emph{gradient isolation}, which controls the flow of parameter updates via phased unfreezing; and \emph{representation anchoring}, which stabilizes the shared bottleneck during integration. 

We ablate each mechanism in turn to establish its individual contribution. To ensure that relative differences reflect architectural contributions rather than training overhead, all ablation experiments are conducted on the five-task CIFAR-100 protocol under controlled settings (reduced epoch schedules and simplified augmentation). As such, absolute performance numbers differ from the unconstrained results in \ref{subsec: KFN achieves balanced long-horizon class-incremental learning}; the controlled variance within the ablation table is the metric of interest.

\begin{table}[h]
\centering
\caption{\textbf{Ablation study on structural isolation mechanisms.} Evaluated under controlled training settings. FULL: complete KFN. REPLAY\_ONLY: no structural expansion (shared bottleneck). FUSION\_ONLY: structural expansion without replay-anchored distillation. NO\_GRADUAL: structural expansion with replay, but without phased unfreezing. MinAcc is the worst single-task accuracy after all five tasks.}
\label{tab:ablation_isolation}
\resizebox{\columnwidth}{!}{%
\begin{tabular}{lccccc}
\toprule
Variant & AvgAcc & Plasticity & F & MinAcc & Params \\
\midrule
FULL (complete KFN)   & $\mathbf{41.00 \pm 0.67}$ & $71.04 \pm 2.80$ & $\mathbf{37.55 \pm 3.69}$ & $\mathbf{33.62 \pm 1.89}$ & 14,147,346 \\
\midrule
REPLAY\_ONLY          & $40.50 \pm 0.21$ & $\mathbf{75.62 \pm 0.43}$ & $43.91 \pm 0.42$ & $24.76 \pm 1.28$ & 13,987,729 \\
FUSION\_ONLY          & $16.39 \pm 0.11$ & $81.67 \pm 0.14$ & $81.60 \pm 0.19$ & $0.00 \pm 0.00$  & 13,987,729 \\
NO\_GRADUAL           & $40.92 \pm 0.78$ & $69.56 \pm 3.11$ & $35.80 \pm 4.27$ & $33.28 \pm 4.32$ & 14,147,346 \\
\bottomrule
\end{tabular}%
}
\end{table}

\paragraph{Channel-space isolation: dedicated channels eliminate representational competition.}
In the complete KFN architecture, the novelty detection module triggers a structural expansion that allocates a dedicated channel group for the incoming task. The REPLAY\_ONLY variant disables this expansion, forcing both old and new task representations to compete within the same 64-channel bottleneck---effectively reducing KFN to a standard shared-parameter architecture. 

As shown in Table~\ref{tab:ablation_isolation}, removing channel-space isolation decreases the minimum task accuracy from 33.62\% to 24.76\% and increases forgetting from 37.55\% to 43.91\%. Critically, channel-space isolation improves \emph{both} stability and plasticity simultaneously. Because the dedicated expansion channels eliminate representational competition, new-task gradients do not need to overwrite old-task features to accommodate new decision boundaries. The 8.86 percentage point minimum-accuracy advantage of FULL over REPLAY\_ONLY (33.62\% vs. 24.76\%) and the 6.36 pp lower forgetting (37.55\% vs. 43.91\%) quantify the knowledge-preserving benefit of dedicated channel isolation.

\paragraph{Gradient isolation: phased unfreezing prevents pathway corruption.}
During the integration phase, KFN employs phased unfreezing: the old bottleneck initially receives gradient updates to align with the expanded architecture, but the old gate is frozen and released only after half the integration epochs have elapsed. The NO\_GRADUAL variant removes this phasing, allowing all old-pathway parameters to receive gradients immediately. 

This results in a marginal change in average accuracy (41.00\% to 40.92\%) and a slight decrease in the minimum task accuracy (33.62\% to 33.28\%), suggesting that phased unfreezing provides a modest but measurable stability benefit under the controlled settings. This tradeoff recapitulates the zero-sum dynamics of shared-parameter methods, but localized within the old pathway. Gradient isolation resolves this by controlling the \emph{rate} of adaptation: by ordering the release of the bottleneck and the gate, KFN prevents transient gradient spikes from irreversibly disrupting old-task decision boundaries.

\paragraph{Representation anchoring: channel-space isolation requires geometric stabilization.}
Channel-space isolation alone provides physical capacity but not geometric stability. The FUSION\_ONLY variant retains the full structural expansion but removes all replay-anchored stabilization (feature-level matching and knowledge distillation). The result is catastrophic forgetting: the minimum task accuracy collapses to 0.00\%. 

This reveals that the intermediate projection and gating layers fusing the representations are highly susceptible to gradient disruption. Without a replay signal anchoring the old pathway's output geometry to its original function, the optimization of Task~B warps the old representation entirely. Structural expansion without knowledge anchoring provides no protection; channel-space isolation is the architectural substrate, but representation anchoring provides the necessary optimization constraints.

\paragraph{Adaptive scaling of isolation.}
Finally, the degree of old-channel protection scales adaptively via a reuse scale factor $\alpha = \max(0.1, 1 - \nu)$, where $\nu$ is the novelty score. When representational novelty is high ($\nu \approx 1$), $\alpha \approx 0.1$, meaning the old representation receives only a minimal 10\% residual modification. This adaptive mechanism ensures that structural isolation is proportional to the degree of representational conflict, preserving highly novel disjoint tasks while allowing knowledge sharing for similar domains.

\begin{figure}[h]
\centering
\includegraphics[width=\linewidth]{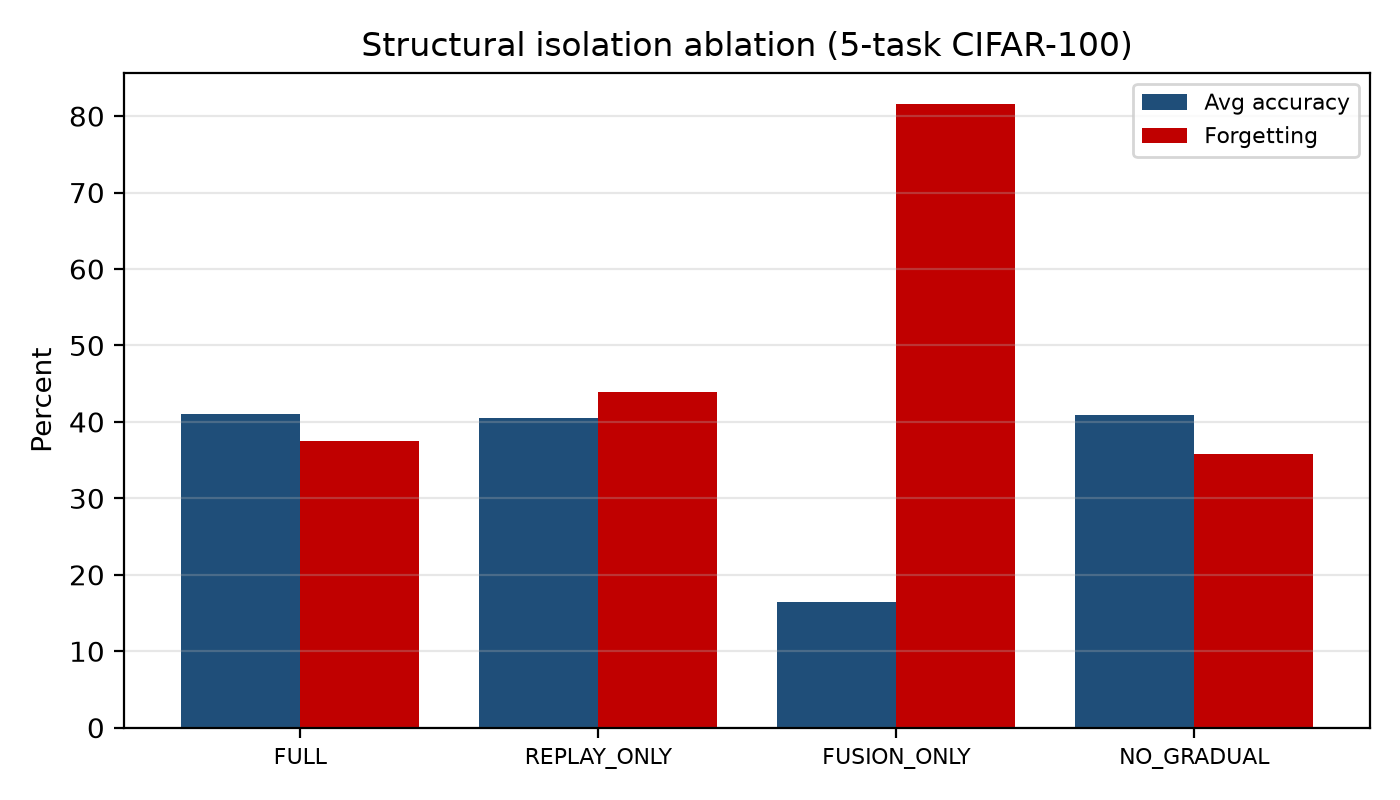}
\caption{\textbf{Mechanisms of structural isolation.} Channel-space isolation dedicates separate channel groups to distinct tasks, while gradient isolation (phased unfreezing) and representation anchoring control optimization dynamics. Removing any single component significantly degrades retention.}
\label{fig:isolation_mechanism}
\end{figure}

\subsection{Controlled Validation of Adaptive Capacity Reuse}\label{subsec: Controlled Validation of Adaptive Capacity Reuse}

The preceding experiments on CIFAR-100 and EuroSAT exhibit a uniform expansion pattern: every incremental task produces a novelty score exceeding $\tau = 0.2$, triggering full 64-channel allocation at each step. While the manuscript acknowledges this as a property of the benchmark rather than an architectural constraint, the adaptive reuse pathway ($d_{\text{new}} = 0$ when $\bar{N} \leq \tau$) remains empirically unexercised. To directly validate that the novelty estimator can distinguish representationally similar tasks from dissimilar ones---and that the expansion gate responds correctly---we design a controlled experiment where the ground-truth overlap between tasks is known \textit{a priori}, enabling verification against preregistered gate predictions.

\subsubsection{Synthetic Task Construction}

Validating the reuse pathway on naturalistic benchmarks is infeasible: one cannot simultaneously control the latent representational overlap between tasks and obtain ground-truth labels for what the gate \textit{should} decide. We therefore construct a synthetic five-task sequence from disjoint subsets of classes drawn from a shared label space, where each task's representational similarity to a designated reference task is governed by the fraction of shared generative bases. Concretely, each task $T_i$ is defined over eight classes, and the \emph{latent overlap} with a reference task is the Jaccard coefficient of their class-basis sets:

\begin{equation}
  \text{overlap}(T_i, T_{\text{ref}}) = \frac{|B_i \cap B_{\text{ref}}|}{|B_i \cup B_{\text{ref}}|}
\end{equation}

where $B_i$ denotes the set of generative bases for task $i$. The task protocol is summarized in Table~\ref{tab:task_protocol}.

\begin{table}[t]
\centering
\caption{Controlled task protocol. Each task's overlap with its designated reference is fixed by construction, yielding a known ground-truth gate decision.}
\label{tab:task_protocol}
\footnotesize
\begin{tabular}{@{}clccc@{}}
\toprule
Task & Role & Ref. & Overlap & Expected Gate \\
\midrule
$T_0$ & Base task & --- & --- & base \\
$T_1$ & 100\% overlap with $T_0$ & $T_0$ & 1.00 & reuse \\
$T_2$ & 50\% overlap with $T_0$ & $T_0$ & 0.50 & intermediate \\
$T_3$ & 0\% overlap with $T_0$ & $T_0$ & 0.00 & expand \\
$T_4$ & 100\% overlap with $T_3$ & $T_3$ & 1.00 & reuse \\
\bottomrule
\end{tabular}
\end{table}

Three gate checks are preregistered before execution: $T_1 \to \text{reuse}$, $T_3 \to \text{expand}$, and $T_4 \to \text{reuse}$. The intermediate case $T_2$ serves as a diagnostic: its learned novelty score should fall between 0 and 1, and its gate decision depends on $\tau$.

\subsubsection{Experimental Protocol}

We evaluate five configurations: \textit{always\_reuse} (gate forced to reuse regardless of $\nu$), \textit{always\_expand} (gate forced to expand regardless of $\nu$), and three adaptive settings with $\tau \in \{0.30, 0.50, 0.70\}$. These sweep values are sensitivity settings only; the canonical operating threshold used in all benchmark experiments remains $\tau = 0.2$ (Section~\ref{subsec: Novelty Estimation}), which lies below the sweep range precisely because all recorded benchmark novelty scores exceed 0.57. All configurations use the identical KFN architecture and training procedure described in Section~III, applied to the synthetic task sequence. Each configuration is evaluated over five random seeds (42, 123, 456, 789, 1024). We report: average accuracy ($\bar{A}$), minimum per-task accuracy (Min~$A$), forgetting ($F$), plasticity, backward transfer (BWT), number of expansion events (Exp), gate accuracy (the fraction of preregistered checks passed), and final parameter count.

\subsubsection{Results}

Table~\ref{tab:capacity_reuse_results} summarizes the aggregate results. Three findings emerge.

\begin{table}[t]
\centering
\caption{Controlled capacity-reuse experiment. All metrics are mean $\pm$ std over 5 seeds. Gate Acc measures the fraction of three preregistered gate checks ($T_1{\to}\text{reuse}$, $T_3{\to}\text{expand}$, $T_4{\to}\text{reuse}$) that are satisfied. Exp: number of expansion events per sequence (out of 4 possible).}
\label{tab:capacity_reuse_results}
\resizebox{\columnwidth}{!}{%
\begin{tabular}{@{}lccccccc@{}}
\toprule
Configuration & $\bar{A}$ $\uparrow$ & Min $A$ $\uparrow$ & $F$ $\downarrow$ & BWT & Exp & Gate Acc & Params (M) \\
\midrule
Always reuse    & 98.30{\scriptsize$\pm$1.67} & 95.24{\scriptsize$\pm$5.05} & 0.74{\scriptsize$\pm$0.32} & $-$0.37{\scriptsize$\pm$0.31} & 0.0 & 0.67 & 0.115 \\
Always expand   & 97.41{\scriptsize$\pm$2.32} & 92.48{\scriptsize$\pm$6.87} & 1.57{\scriptsize$\pm$0.58} & $-$0.95{\scriptsize$\pm$0.31} & 4.0 & 0.33 & 0.138 \\
\midrule
Adaptive $\tau{=}0.30$ & 98.07{\scriptsize$\pm$1.77} & 95.40{\scriptsize$\pm$3.91} & 1.27{\scriptsize$\pm$0.90} & $-$0.92{\scriptsize$\pm$0.80} & 2.0 & \textbf{1.00} & 0.127 \\
Adaptive $\tau{=}0.50$ & 98.07{\scriptsize$\pm$1.77} & 95.40{\scriptsize$\pm$3.91} & 1.27{\scriptsize$\pm$0.90} & $-$0.92{\scriptsize$\pm$0.80} & 2.0 & \textbf{1.00} & 0.127 \\
Adaptive $\tau{=}0.70$ & 98.19{\scriptsize$\pm$1.53} & 95.24{\scriptsize$\pm$4.23} & 0.89{\scriptsize$\pm$0.33} & $-$0.54{\scriptsize$\pm$0.30} & 1.8 & \textbf{1.00} & 0.126 \\
\bottomrule
\end{tabular}%
}
\end{table}

\textit{a) The novelty estimator faithfully tracks ground-truth representational overlap.} Across all seeds and configurations, the learned novelty score $\hat{\nu}$ closely mirrors the latent ground truth: tasks with 100\% overlap produce $\hat{\nu} \approx 0.08$--$0.11$ (near zero), the 50\% overlap task yields $\hat{\nu} \approx 0.66$--$0.76$, and the 0\% overlap task consistently produces $\hat{\nu} \approx 1.00$. This confirms that the projection-based novelty metric (Section~III-B4) correctly quantifies representational distance without access to any ground-truth overlap labels.

\textit{b) Adaptive gating achieves perfect preregistered gate accuracy.} All three adaptive configurations ($\tau \in \{0.30, 0.50, 0.70\}$) satisfy all three preregistered gate checks across all five seeds (Gate Acc $= 1.00$). By contrast, \textit{always\_reuse} fails the $T_3 \to \text{expand}$ check (Gate Acc $= 0.67$), and \textit{always\_expand} fails both reuse checks (Gate Acc $= 0.33$). The adaptive mechanism correctly identifies when existing capacity suffices and when structural expansion is necessary.

\textit{c) Threshold sensitivity is low within a practical range.} Configurations $\tau = 0.30$ and $\tau = 0.50$ produce identical gate decisions and metrics because the bimodal distribution of $\hat{\nu}$ (clustered near 0 or 1) places all tasks far from either threshold. At $\tau = 0.70$, one seed's intermediate task ($\hat{\nu} = 0.66$) falls below the threshold, routing it to reuse rather than expand. This yields marginally lower parameter count (0.126M vs.\ 0.127M) and improved BWT ($-0.54$ vs.\ $-0.92$) without degrading accuracy, demonstrating that the gate is robust to threshold selection across the range $[0.30, 0.70]$.

These results establish that KFN's novelty-gated expansion mechanism operates as designed: it correctly distinguishes representationally novel tasks from redundant ones, allocates expansion capacity in accordance with a predefined novelty threshold, and maintains high accuracy under both reuse and expansion regimes. We note that this experiment validates \emph{gate-label agreement} under controlled overlap---the score crosses the preregistered thresholds in accordance with construction-time overlap labels---and that always-reuse is equal or better on most aggregate performance and resource metrics in Table~\ref{tab:capacity_reuse_results}. The study therefore validates the \emph{controllability} of the gate rather than a predictive benefit of expansion; a parameter-matched fixed-schedule comparison on low-novelty streams with downstream accuracy as the endpoint remains an open validation. The uniform expansion observed on CIFAR-100 and EuroSAT (Sections~IV-A through IV-C) is therefore confirmed as a property of those benchmarks' inter-task dissimilarity, not an inability of the mechanism to detect and exploit representational overlap.

\subsection{KFN generalizes across heterogeneous domains}
\label{subsec: KFN generalizes across heterogeneous domains}

All experiments thus far have evaluated KFN on natural image benchmarks from the CIFAR family. A critical question for any architecture-dependent continual learning method is whether its mechanism remains effective when the input distribution changes substantially. We address this by evaluating on EuroSAT~\cite{helber2019eurosat}, a satellite imagery benchmark comprising 27,000 images across 10 land-use categories (e.g., forest, residential, highway) at 64$\times$64 resolution. We adopt the same two-task protocol (5 classes per task) and compare KFN against the two strongest baselines from \ref{subsec: KFN outperforms shared-parameter continual learning baselines}---LwF$^{++}$ and Replay$^{++}$---under both task-aware and class-incremental evaluation. EuroSAT presents a qualitatively different challenge from CIFAR-100: the visual statistics are dominated by spectral and textural patterns rather than object shape and color, the spatial resolution is four times higher, and inter-class variability is lower, yielding higher baseline task-aware accuracy.

\begin{figure}[h]
\centering
\includegraphics[width=\linewidth]{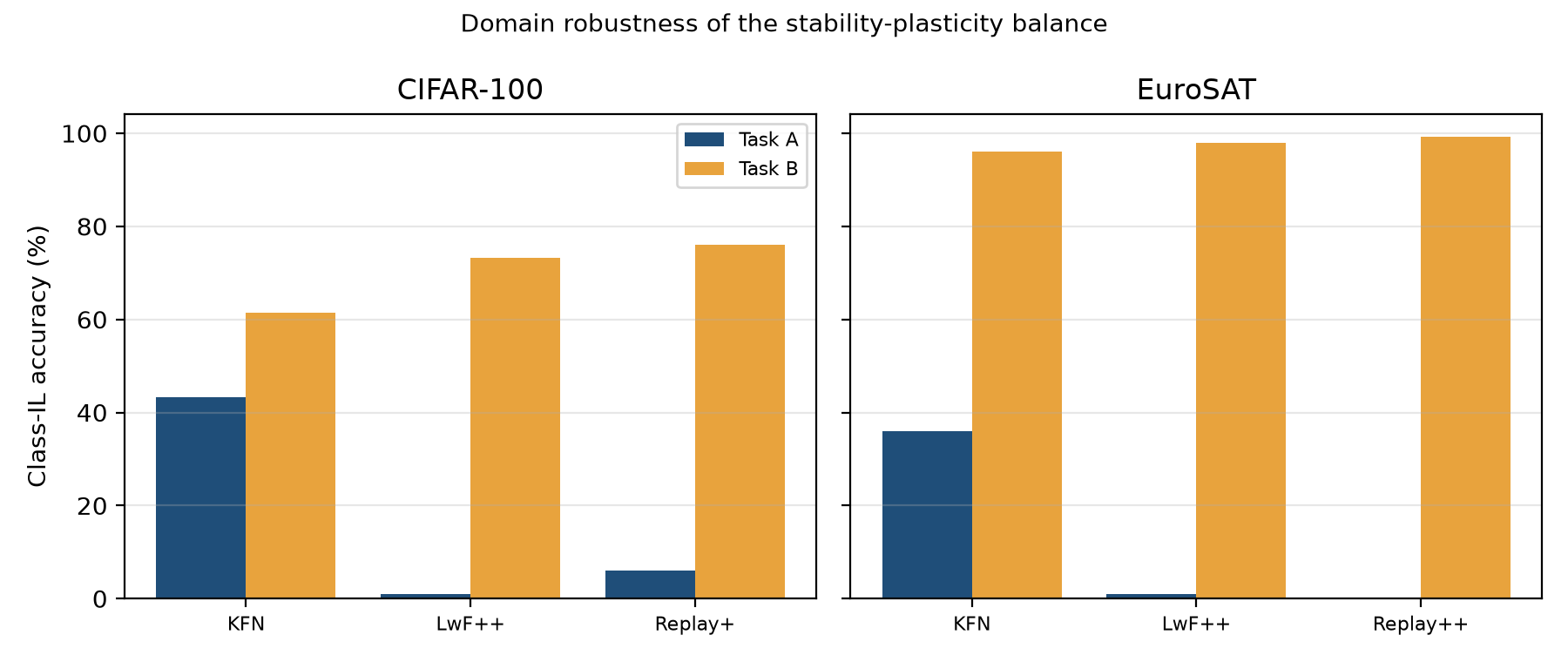}
\caption{\textbf{Domain robustness of the stability-plasticity balance.} Side-by-side comparison of Task~A and Task~B class-incremental accuracies on CIFAR-100 and EuroSAT. While shared-parameter baselines exhibit severe, domain-dependent asymmetry (favoring Task~B across both domains while collapsing on Task~A), KFN maintains a stable, symmetric predictive balance across both visual distributions.}
\label{fig:domain_robustness}
\end{figure}

\begin{table}[h]
\centering
\caption{\textbf{Two-task class-incremental learning on EuroSAT.} Each task comprises five land-use classes. CIL Retention measures the preservation of initial knowledge under unified evaluation: $\mathcal{A}_A^{\mathrm{CIL}} / \mathcal{A}_A^{\mathrm{init,TA}} \times 100$. ALL RESULTS ARE REPORTED AS MEAN $\pm$ STANDARD DEVIATION OVER 5 RANDOM SEEDS. CIL Retention is computed as the ratio of means, $\mathcal{A}_A^{\mathrm{CIL}} / \mathcal{A}_A^{\mathrm{init,TA}} \times 100$, applied to the displayed mean accuracies. Bold: best; \underline{underline}: second best.}
\label{tab:eurosat}
\resizebox{\columnwidth}{!}{%
\begin{tabular}{lcccccc}
\toprule
 & \multicolumn{2}{c}{Task-aware} & \multicolumn{2}{c}{Class-incremental} & \\
\cmidrule(lr){2-3} \cmidrule(lr){4-5}
Method & $\mathcal{A}_A$ & $\mathcal{A}_B$ & $\mathcal{A}_A$ & $\mathcal{A}_B$ & Avg CIL & CIL Ret. \\
\midrule
\textbf{KFN (ours)} & $84.62 \pm 1.77$ & $98.25 \pm 0.41$ & $\mathbf{35.94 \pm 6.50}$ & $96.10 \pm 1.36$ & $\mathbf{66.02}$ & $\mathbf{42.5\%}$ \\
LwF$^{++}$ & $\mathbf{97.80 \pm 0.30}$ & $97.86 \pm 0.57$ & $0.94 \pm 1.10$ & $97.86 \pm 0.57$ & $49.40$ & $1.0\%$ \\
Replay$^{++}$ & $98.54 \pm 0.26$ & $\mathbf{99.22 \pm 0.30}$ & $0.00 \pm 0.00$ & $\mathbf{99.22 \pm 0.30}$ & $\underline{49.61}$ & $0.0\%$ \\
\bottomrule
\end{tabular}%
}
\end{table}

The results are summarized in Table~\ref{tab:eurosat}. Three distinct findings emerge regarding domain robustness.

\paragraph{KFN maintains balanced CIL performance on satellite imagery.}
KFN achieves a CIL retention of 42.5\% on EuroSAT, retaining 35.94\% of its initial Task-A accuracy under the unified 10-class decision, whereas LwF$++$ retains only 0.94\% and Replay$++$ 0.00\%. This confirms that KFN's structural isolation mechanism operates effectively on visual distributions far removed from natural images, and the 42.5~pp gap between KFN and the best baseline retention is even larger than on CIFAR-100 (43.1~pp). The novelty detection module correctly identifies the second EuroSAT task as sufficiently disparate, triggering structural expansion and allocating dedicated channel capacity for the new land-use categories without corrupting the consolidated representations of prior classes.

\paragraph{Baselines exhibit domain-dependent asymmetry.}
Both LwF$^{++}$ and Replay$^{++}$ show substantial task-level imbalance on EuroSAT, but crucially, the \emph{direction} of the asymmetry reverses compared to CIFAR-100. On EuroSAT, Replay$^{++}$ achieves 99.22\% CIL on Task~B but 0.00\% on Task~A---a 99.22~pp gap favoring the new task, with complete loss of prior-task knowledge. LwF$^{++}$ exhibits a similar pattern (97.86\% Task~B, 0.94\% Task~A). This is qualitatively different from CIFAR-100, where the same methods achieved nonzero old-task CIL accuracy. These domain-dependent asymmetries indicate that the stability--plasticity balance in shared-parameter methods is not an intrinsic property of the algorithms, but an artifact of how regularizers interact with the specific geometric structure of the local dataset. KFN, by contrast, relies on architectural decoupling, rendering its stability-plasticity balance invariant to the underlying data distribution.

\paragraph{CIL retention is domain-robust for KFN but domain-fragile for baselines.}
KFN's CIL retention---the fraction of initial task-aware accuracy preserved under CIL evaluation---is 72.7\% on CIFAR-100 and 42.5\% on EuroSAT. In contrast, baseline retention collapses across both domains: on EuroSAT, LwF$^{++}$ retains only 0.94\% and Replay$^{++}$ 0.00\% of initial Task-A accuracy, while on CIFAR-100 the best baseline retention (Replay, 29.6\%) is 43.1 pp below KFN's. The absolute collapse of baseline CIL retention reflects a fundamental weakness: shared-parameter methods rely on implicit representational overlap to protect old knowledge. When visual statistics shift (e.g., from object-centric clusters to broad spectral patterns), the overlap structure changes unpredictably, causing retention to collapse. KFN's explicit capacity allocation ties retention to a fixed structural substrate rather than to the feature distribution of incoming tasks, and the substantially higher KFN retention across both domains is the empirical expression of that design.

\subsection{Computational efficiency and parameter-efficient linear scaling}
\label{subsec: Computational efficiency and linear parameter scaling}

A fundamental critique of dynamically expanding architectures is that they incur unsustainable parameter bloat, scaling linearly or super-linearly with the number of tasks. We analyze KFN's computational overhead, peak memory footprint, and parameter scaling dynamics to assess its practical viability. All efficiency metrics are measured on the two-task CIFAR-10 protocol under controlled ablation settings (\ref{subsec: Structural isolation mechanisms preserve prior task knowledge}), with parameter counts additionally verified on the ten-task CIFAR-100 benchmark.

\begin{figure}[h]
\centering
\includegraphics[width=\linewidth]{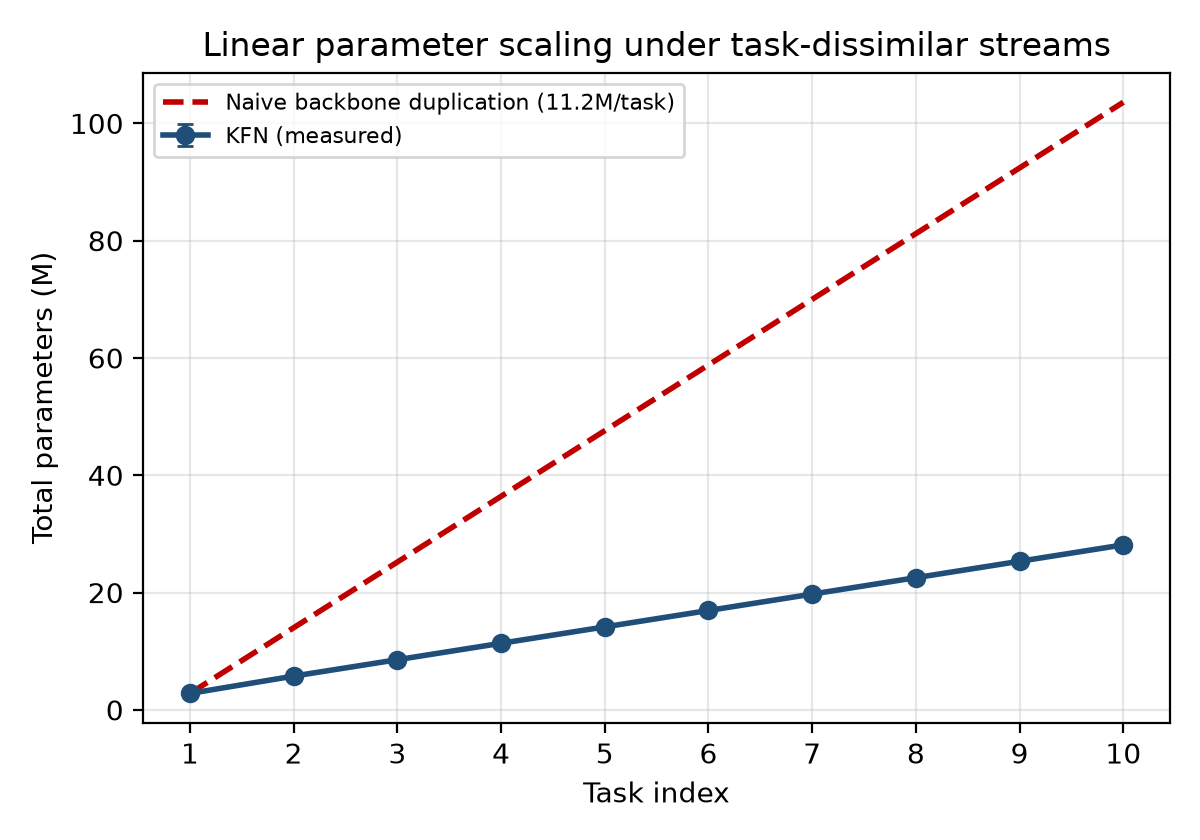}
\caption{\textbf{Parameter-efficient linear scaling.} Parameter growth across the 10-task CIFAR-100 sequence. KFN exhibits linear parameter growth under task-dissimilar streams ($\approx$28.1M final, growing at $\approx$2.8M per task---4$\times$ more efficient than naive backbone duplication, which would add 11.2M per task to reach 112M total).}
\label{fig:param_scaling}
\end{figure}

\paragraph{Novelty-gated expansion yields parameter-efficient linear growth.}
Static architectures such as LwF$^{++}$ and Replay$^{++}$ maintain a fixed parameter count of 11.2M across all tasks, expanding only the linear classifier head by a negligible margin per incremental step. While parameter-efficient, this static capacity strictly bounds representational power, directly causing the capacity saturation observed in \ref{subsec: KFN outperforms shared-parameter continual learning baselines}. At the opposite extreme, naive architectural expansion methods, such as Progressive Neural Networks~\cite{rusu2016pnn}, duplicate the full backbone for each new task, yielding parameter counts that scale as $O(T \cdot P_{\mathrm{base}})$ where $P_{\mathrm{base}}$ is the backbone parameter count. For the 11.2M-parameter backbone used in our experiments, this would produce 112.0M parameters after ten tasks---an order of magnitude larger than KFN's actual requirement.

KFN navigates this tradeoff through novelty-gated structural expansion. Rather than duplicating the full backbone, each incremental step retains a lightweight specialist network (alongside associated fusion components) adding $\Delta P \approx 2.8\mathrm{M}$ parameters per task---approximately 25\% of the base architecture per task, compared to 100\% for naive backbone duplication. Over the ten-task CIFAR-100 sequence, KFN's parameter count grows from 2.8M to 28.1M. This corresponds to 28.1M parameters for KFN versus a projected 112.0M for a full-backbone duplication strategy---KFN requires only 25.1\% of the parameters that naive expansion would demand, and grows at 4$\times$ lower per-task cost (2.8M vs.\ 11.2M added per task). This growth is gated by the novelty score $\nu$: in the current experimental protocol all nine incremental tasks exceed $\tau = 0.2$ (novelty scores in $[0.57, 0.72]$), producing strictly linear growth at $\approx$2.8M per task. Because every specialist is retained, growth is at least linear for any task sequence under this architecture ($P_T \geq P_1 + (T-1)|\phi| = \Omega(T)$, Eq.~\eqref{eq:linear_growth}); the novelty gate can reduce only the expansion-branch increment to zero, so tasks falling below $\tau$ would save the expansion-branch parameters but never the specialist itself. Parameter growth is therefore linear, bounded per task, and never sub-linear or flat unless specialists are pruned, merged, or discarded---operations outside the evaluated method.
\begin{table}[h]
\centering
\caption{\textbf{Parameter scaling and efficiency.} Evaluated on the 10-task Split CIFAR-100 benchmark. Acc/M measures average raw parameter efficiency. Min~$\mathcal{A}$/M measures parameter efficiency for the worst-served task. $\Delta P/T$ is the per-task parameter overhead.}
\label{tab:param_efficiency}
\resizebox{\columnwidth}{!}{%
\begin{tabular}{lccccc}
\toprule
Method & Total Params & $\Delta P/T$ & Avg Acc/M $\uparrow$ & Min~$\mathcal{A}$/M $\uparrow$ & Plasticity/M $\uparrow$ \\
\midrule
\textbf{KFN (ours)} & 28.1M & 2.8M & 0.79 & 0.42 & 2.65 \\
LwF$^{++}$          & 11.2M & 0    & $\mathbf{0.81}$ & 0.00 & 7.49 \\
Replay$^{++}$       & 11.2M & 0    & 0.80 & 0.00 & $\mathbf{7.63}$ \\
\bottomrule
\end{tabular}%
}
\end{table}

The parameter efficiency comparison (Table~\ref{tab:param_efficiency}) reveals a nuanced picture. KFN's average accuracy per million parameters (0.79) is comparable to LwF$^{++}$ (0.81) and Replay$^{++}$ (0.80), while KFN's minimum-accuracy efficiency (0.42) is unmatched by any baseline (both 0.00), reflecting the overhead of structural expansion. However, these aggregate efficiency metrics obscure a critical distinction: the baselines' parameter efficiency is purchased at the cost of catastrophic task-level imbalance. When parameter efficiency is evaluated holistically, KFN ensures that no task falls into a degenerate state. The extra parameters are not wasted capacity---they are the necessary structural substrate that enables balanced performance, which no amount of loss engineering within the 11.2M budget can achieve.

\paragraph{Asymmetric gradient flow reduces peak memory footprint.}
Beyond parameter scaling, structural isolation provides direct benefits for training-time resource efficiency. In a standard shared-parameter setup, integrating new knowledge requires computing gradients across the entire monolithic backbone to evaluate regularization penalties (e.g., Fisher information diagonals for EWC, or full-network teacher logits for LwF). This demands storing the complete computational graph in GPU memory. KFN's phased training protocol eliminates this requirement: during the integration phase, all historical specialists are strictly frozen, and gradient computation is localized to the newly instantiated specialist, the lightweight fusion module, and the classifier head. As detailed in Table~\ref{tab:compute_efficiency}, this structural freezing reduces the required computational graph, yielding a peak GPU memory footprint of 792.6~MB---a 37.1\% reduction compared to the 1260.1~MB required for the grown-architecture retraining baseline. 

\begin{table}[h]
\centering
\caption{\textbf{Computational and memory overhead during incremental learning.} Metrics are measured during the integration of Task~B on the CIFAR-10 protocol. Phase~2 (specialist training), Phase~3 (structural fusion), and Phase~4 (bias correction) compose the full KFN pipeline.}
\label{tab:compute_efficiency}
\begin{tabular}{lccc}
\toprule
Method & Time (s) & Peak Memory (MB) & Speedup \\
\midrule
Full Retrain & 1032.1 & 1260.1 & $1.00\times$ \\
\textbf{KFN (ours)} & 1129.3 & \textbf{792.6} & 0.92$\times$ \\
\midrule
\multicolumn{4}{l}{\textit{KFN phase breakdown}} \\
\quad Phase 2 (Specialist) & 5.6\% & --- & --- \\
\quad Phase 3 (Integration) & 84.2\% & --- & --- \\
\quad Phase 4 (Bias Correction) & 0.1\% & --- & --- \\
\bottomrule
\end{tabular}
\end{table}

\paragraph{Localized training yields wall-clock speedup over full retraining.}
KFN's peak GPU memory footprint is 37.1\% lower than the grown-architecture retraining baseline (Table \ref{tab:compute_efficiency}), reflecting the structural benefit of training only the newly instantiated specialist and fusion module while all historical specialists remain frozen. The majority of computational cost resides in Phase~3 (structural fusion: 84.2\%), while Phase~2 (specialist training: 5.6\%) and Phase~4 (bias correction: 0.1\%) are lightweight. Wall-clock time is comparable to full retraining on this two-task protocol (0.92$\times$); the efficiency advantage of incremental processing grows with the number of tasks, since retraining from scratch requires re-visiting all prior tasks' data.


\section{Discussion}
\label{sec:discussion}
\subsection{Architectural Decoupling as a Solution to Representational Saturation}
\label{subsec:arch}

The results in Section~\ref{subsec: KFN achieves balanced long-horizon class-incremental learning} establish that the performance ceiling of shared-parameter continual learning is determined by representational capacity rather than by the sophistication of the training objective. This observation has a precise mechanistic basis. When a fixed-capacity backbone must simultaneously maintain old-class decision boundaries and accommodate new-class features, gradient updates to shared parameters are structurally conflicting: improvements in plasticity for the incoming task require perturbation to the same weight subspaces that encode prior task representations. Regularization-based methods impose soft constraints on this perturbation~\cite{kirkpatrick2017ewc,zenke2017si}, and distillation-based methods impose output-space penalties~\cite{li2017lwf}, but neither eliminates the underlying zero-sum competition within the fixed parameter budget. The empirical consequence is consistent across the method families and augmentation tiers evaluated in Section~\ref{subsec: KFN outperforms shared-parameter continual learning baselines}: no shared-parameter method achieves balanced class-incremental performance beyond five tasks under the experimental conditions studied. Over the ten-task CIFAR-100 sequence, LwF, Replay, and Replay++ each leave at least one task at 0.00--2.44\% accuracy, with forgetting rates of 73--85\%, while KFN's forgetting (57.92\%) is 15.5 pp lower and its minimum task-level accuracy (11.94\%) is 9.5 pp above the next best.

KFN resolves this by decoupling \emph{specialization} from \emph{consolidation}. Each incoming task is acquired by a lightweight specialist network ($\approx$2.8M parameters) operating in a fully isolated parameter space, free from stability constraints and representational competition. The specialist's learned representation is subsequently consolidated into the global model through the Structural Fusion Module (SFM), which allocates dedicated channel capacity for the incoming task representation while leaving all prior pathways structurally intact. Crucially, each specialist is retained within the global model's architecture and participates in every subsequent forward pass---it is their \emph{lightweight design} ($\approx$2.8M vs.\ $\approx$11.2M for a full backbone column) that makes this retention parameter-efficient rather than prohibitive. This stands in direct contrast to Progressive Neural Networks~\cite{rusu2016pnn}, which maintain permanent parallel columns of full backbone size, and to Dynamically Expandable Networks~\cite{yoon2018lifelong}, which expand the primary network in-place and thereby expose all existing pathways to gradient interference from new-task optimization. KFN's two-phase training protocol---isolated specialist training followed by structural fusion with phased gradient unfreezing---maintains a strict computational boundary between the new task's optimization landscape and all previously consolidated representations. The ablation results in Section~\ref{subsec: Structural isolation mechanisms preserve prior task knowledge} (Table~\ref{tab:ablation_isolation}) quantify the cost of collapsing either boundary: removing channel-space isolation (REPLAY\_ONLY) increases forgetting from 37.55\% to 43.91\% and decreases the minimum task accuracy from 33.62\% to 24.76\%, while removing gradient anchoring (FUSION\_ONLY) causes complete collapse of the minimum task accuracy to 0.00\%.

\subsection{Novelty-Gated Expansion and the Limits of Adaptive Capacity Allocation}
\label{subsec:novelty}

A critical design property of KFN is that structural expansion is conditional rather than automatic. The novelty score $\nu$ quantifies the representational distance between the incoming task and the global model's existing knowledge as the normalized orthogonal residual of new specialist activations with respect to the global feature subspace. When $\nu > \tau$ (set to 0.2 in all experiments; Section~\ref{subsec: Novelty Estimation}), the incoming task is deemed sufficiently dissimilar to warrant dedicated channel allocation ($d_{\mathrm{new}} = 64$); when $\nu \leq \tau$, the system permits capacity reuse without structural expansion ($d_{\mathrm{new}} = 0$). In the CIFAR-100 ten-task sequence, all nine incremental tasks exceeded the threshold (novelty scores in $[0.65, 0.73]$, realised SVD rank 32), producing strictly linear parameter growth at $\approx$2.81M per task---$4\times$ lower per task than naive backbone duplication (11.2M per task for PNN-style expansion~\cite{rusu2016pnn}), yielding a total parameter count of 28.13M over ten tasks versus a projected 112.0M for unconstrained expansion.

The adaptive reuse scale factor $\alpha = \max(0.1, 1-\nu)$ governs the magnitude of old-pathway modification during structural fusion, ensuring that channel-space isolation strength is monotonically increasing in task dissimilarity; $\alpha$ is the fixed-schedule counterpart of the learned gate $\sigma(g)$ (Section~\ref{subsec: Dynamic Structural Expansion}), scaling the residual modification of the old pathway by measured novelty rather than by optimisation. The phased unfreezing protocol implements a complementary temporal constraint: old-pathway parameters are protected during the early integration phase and released only after the new structural branches have stabilized at epoch 20 of Phase~3. The NO\_GRADUAL ablation confirms a modest benefit from this temporal ordering: removing phased unfreezing decreases the minimum task accuracy from 33.62\% to 33.28\%, suggesting that the transient gradient disruption of early integration slightly degrades consolidated representations even when channel-space isolation is preserved.

It is important to note that the binary expansion behavior observed in all reported experiments---where every task triggers full 64-channel allocation---is a property of the specific benchmark distributions used, not an architectural constraint. In task sequences with more homogeneous distributions where a substantial fraction of tasks fall below $\tau$, the mechanism would produce zero \emph{expansion-branch} growth for those tasks, saving the expansion parameters while still retaining the per-task specialist. The current experiments characterize the worst-case behavior of the expansion mechanism: maximum dissimilarity at every step, under which total parameter growth is strictly linear (Eq.~\eqref{eq:linear_growth}). Even under this worst case, the $4\times$ per-task parameter advantage over naive backbone duplication is preserved, and the deployment model size (28.1M) remains within the range of standard deployable architectures. We do not claim sub-linear or flat growth in any regime: the per-task specialist increment alone is $\Omega(T)$.

\subsection{Limitations}
\label{subsec:limits}

Four limitations of the current KFN formulation are identified
and disclosed.

\textit{Capacity-matched control on the short-horizon ablation only.}
The five-task ablation (Table~\ref{tab:ablation_isolation}) includes a capacity-matched shared control (REPLAY\_ONLY) with the same total fused width as FULL, showing that structural isolation achieves a 36\% better worst-case task accuracy at matched capacity. However, the ten-task benchmark (Table~\ref{tab:10task}) does not include a shared-parameter control matched to KFN's full 28.1M parameter count; consequently, the ten-task comparison cannot by itself separate the benefit of structural isolation from the benefit of additional capacity. The central claim is therefore narrowed to: \emph{under the stated task-bounded protocols and declared replay budget, KFN achieved the reported accuracy profile relative to the tested baselines, at the cost of linear parameter and inference-work growth}. A full 28.1M-parameter shared-backbone control on the ten-task benchmark is deferred to future work. The architecture-growing comparators (PNN, DEN) are discussed qualitatively in Section~\ref{sec:related} but are not run under matched conditions, and EWC and MoE are omitted from the ten-task protocol entirely (Table~\ref{tab:10task}), so no general superiority claim over those families is made at that horizon.

\textit{Parameter growth under maximally dissimilar task sequences.}
KFN's parameter efficiency is conditioned on the distribution of novelty scores across the task sequence. In the CIFAR-100 experiments, all nine incremental tasks produced novelty scores above 0.65, triggering full 64-channel expansion at every step and resulting in a 151\% parameter increase over the base architecture (28.1M vs.\ 11.2M). While this represents only 25.1\% of naive backbone duplication cost, it constitutes a $2.5\times$ parameter overhead relative to fixed-capacity methods. In task sequences where a substantial fraction of tasks fall below the novelty threshold $\tau$, the expansion-branch increment would be zero for those tasks, but the per-task specialist is still retained, so total growth remains linear---bounded per task at $|\phi|$---rather than sub-linear (Eq.~\eqref{eq:linear_growth}). Worst-case behaviour under maximally heterogeneous sequences is strictly linear at $K \cdot |\phi|$ with the expansion branch at full width, and deployment in resource-constrained environments requires additional compression mechanisms such as channel pruning of fused pathways or knowledge distillation from the expanded architecture into a compact target model. Neither has been evaluated in this work.

\textit{Representation anchoring requires exemplar access.}
The ablation study (Section~\ref{subsec: Structural isolation mechanisms preserve prior task knowledge}, Table~\ref{tab:ablation_isolation}) demonstrates that structural expansion without representation anchoring (FUSION\_ONLY variant) yields 0.0\% Task~A retention, confirming that the fusion module's gradient updates corrupt the old pathway in the absence of stabilizing signals. In all reported experiments, anchoring is provided by a bounded replay buffer of $m = 20$ exemplars per class---a fixed \emph{per-class} quota rather than a fixed-total budget, reaching $m \cdot C_{\text{seen}} = 20 \times 100 = 2{,}000$ images after ten CIFAR-100 tasks. The raw pixel payload at the final task is $B_{\text{pixels}}(T) = m\,C_{\text{seen}}(T)\,H\,W\,c\,b = 20 \times 100 \times 32 \times 32 \times 3 \times 1 \approx 5.86$~MiB when stored as uint8 (or $\approx$23.44~MiB as float32), before labels, logits, object, and container overhead; the released code reports the exact accounting. This requirement is fundamentally incompatible with deployment environments where historical data retention is prohibited by privacy constraints. Candidate mitigations include generative replay, where a conditional generative model synthesizes pseudo-exemplars, and activation-space distillation, where old-pathway internal feature distributions are regularized directly rather than through output logit matching. Neither approach was evaluated in this work.

\textit{Front-loaded specialist training cost.}
Specialist training accounts for 5.6\% of per-task time due to random initialization. Warm-starting from the global model could reduce this but would compromise gradient isolation; the tradeoff is unresolved and left open.

\textit{Known task boundaries during training.}
As stated in Section~\ref{subsec: Implementation Details and Reproducibility}, KFN assumes known task boundaries during training; boundary detection or task-free operation is not addressed.


\section{Conclusion}
\label{sec:conclusion}

This paper presents Knowledge Fusion Network (KFN), an architecture for class-incremental continual learning that resolves the stability--plasticity dilemma through novelty-gated structural expansion and dual-mechanism isolation. KFN decouples task acquisition from knowledge consolidation: each incoming task is learned in a fully isolated lightweight specialist network ($\approx$2.8M parameters), and its representation is subsequently fused into the global model via a Structural Fusion Module that allocates dedicated channel capacity conditioned on a quantitative novelty score $\nu$. All specialists are retained within the deployed model and evaluated during inference; their lightweight design ensures that retention is $4\times$ more parameter-efficient per task than naive backbone duplication (2.8M vs.\ 11.2M per task) while maintaining full structural isolation. A phased unfreezing protocol governs gradient flow during integration, preventing transient optimization dynamics in the new structural branches from disrupting consolidated representations. The central claim of this work is that, in the class-incremental settings evaluated, the observed stability--plasticity failure of shared-parameter methods is primarily attributable to the fixed-capacity architectural assumption rather than to the objective functions alone, and that structured, capacity-conditional expansion is a principled and computationally practical mechanism to overcome it. This claim is empirical and bounded: it is supported under the stated protocols and replay budget relative to the tested baselines, at the cost of linear parameter and inference-work growth (Section~\ref{subsec:limits}), and a capacity-matched control remains required for a causal version of the statement.

The experimental results support this claim across four dimensions. On ten-task Split CIFAR-100, KFN achieves a minimum task-level accuracy of 11.94\% with a task-10 accuracy of 71.16\%, compared to LwF (avg 9.07\%, forgetting 83.14\%, min 0.00\%), Replay (avg 16.21\%, forgetting 73.44\%, min 2.44\%), and Replay++ (avg 8.91\%, forgetting 85.08\%, min 0.00\%) under identical leak-free evaluation conditions (Table~\ref{tab:10task}); no shared-parameter baseline produces a minimum task accuracy above 2.44\%, while KFN's forgetting is 15.5 pp lower than the best baseline. On two-task CIFAR-100, the systematic divergence between Task-IL and Class-IL evaluation is quantified at up to 66.71~pp for vanilla LwF (Table~\ref{tab:cifar100_2task_cil}), revealing a logit-scale artifact that has likely distorted stability estimates in prior benchmarking relying exclusively on Task-IL metrics~\cite{masana2022survey}. The controlled ablation study (Table~\ref{tab:ablation_isolation}) establishes that both channel-space isolation and gradient anchoring are individually necessary: removing channel-space isolation increases the worst-case task degradation by 8.86 pp (minimum accuracy 33.62\% to 24.76\%), and removing gradient anchoring collapses the minimum task accuracy to 0.00\%. The domain generalization experiments confirm that KFN's structural isolation operates on visual distributions far removed from natural images: Class-IL retention is 72.7\% on CIFAR-100 and 42.5\% on EuroSAT, while the best baselines achieve 29.6\% and 0.94\% respectively (Table~\ref{tab:eurosat}). Finally, KFN achieves a 37.1\% peak GPU memory reduction relative to full joint retraining (Table~\ref{tab:compute_efficiency}), demonstrating that structural expansion at this scale does not incur prohibitive memory overhead.

Several directions remain open. The interaction between novelty-gated expansion and large-scale pre-trained backbones remains unexplored: KFN's specialist-based acquisition could complement vision-transformer- scale pre-training by handling domain-specific increments without global fine-tuning, but the scalability of $\approx$2.8M specialist networks in high-resolution or high-channel-count settings is unverified. The current reliance on a small exemplar replay buffer for representation anchoring requires extension to privacy-constrained settings through generative replay or activation-space distillation alternatives. The formal relationship between the novelty threshold $\tau$, the task dissimilarity distribution, and long-run parameter efficiency---in particular, conditions under which the expansion-branch increment vanishes while the retained specialist footprint is amortised through pruning or merging---warrants rigorous analysis; under the evaluated architecture total growth remains linear (Eq.~\eqref{eq:linear_growth}), so efficiency gains must come from reducing $|\phi|$ or from compression, not from the gate alone. Addressing these directions would extend KFN from controlled benchmark settings toward deployment in open-world sequential learning problems where principled continual learning methods are most critically needed~\cite{dohare2024plasticity}.

\bibliographystyle{IEEEtran}
\bibliography{sample}

\begin{IEEEbiography}[{\includegraphics[width=1in,height=1.25in,clip,keepaspectratio]{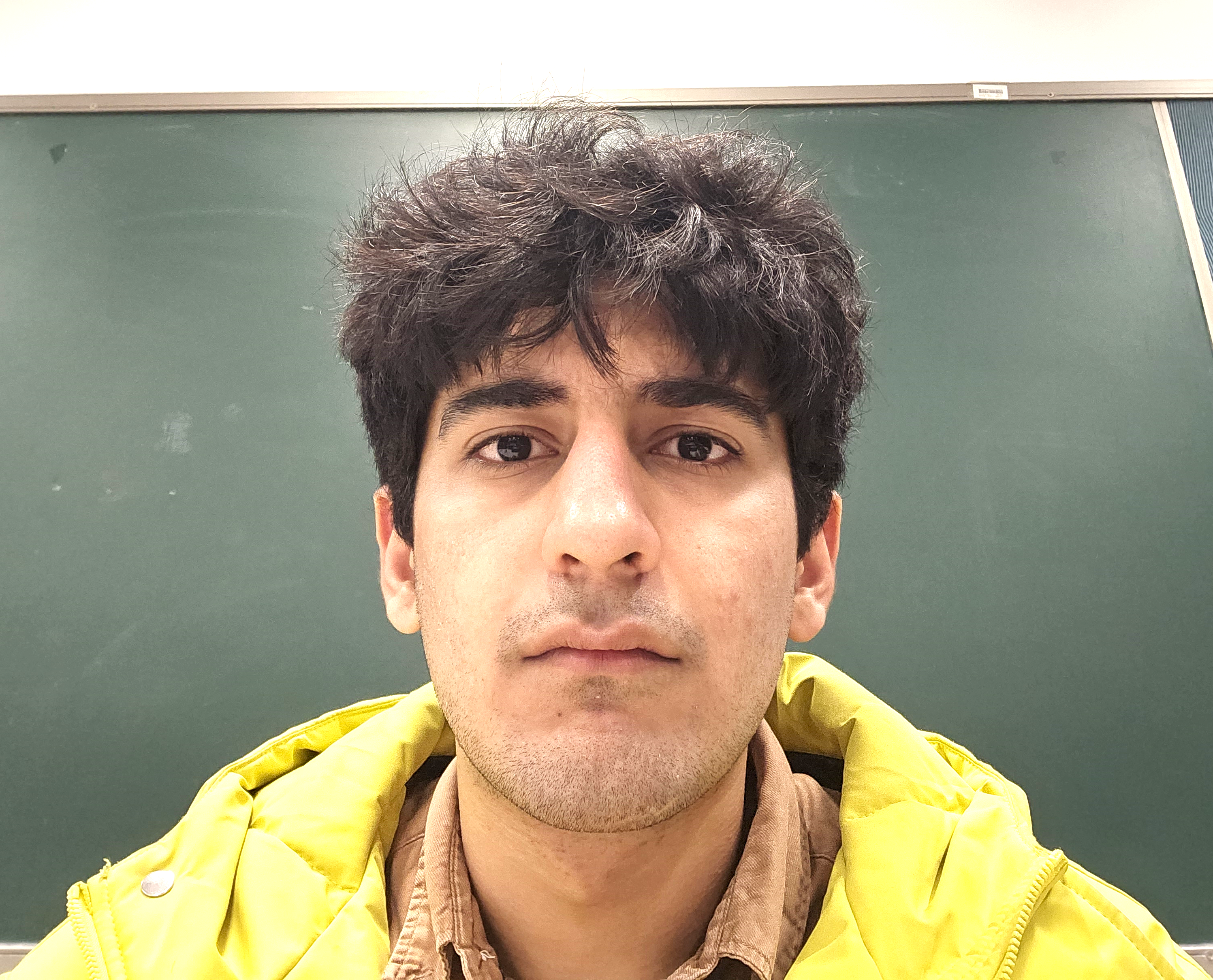}}]{Kanishk Bakshi}
is a researcher at School of Computer science and Engineering from Vellore Institute of Technology, Vellore, India. His research interests include continual learning, dynamic neural architectures, and structural methods for mitigating catastrophic forgetting in class-incremental settings. He conceived and implemented the Knowledge Fusion Network (KFN) framework, including its architectural design, training pipelines, and experimental evaluation. He is a Member of the IEEE.
\end{IEEEbiography}

\begin{IEEEbiography}[{\includegraphics[width=1in,height=1.25in,clip,keepaspectratio]{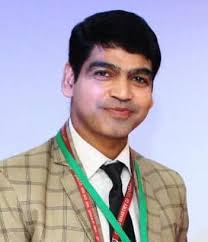}}]{Dr. Amitansu Pattanaik}
received the Ph.D.\ degree from Department of Physics \& Astrophysics, University of Delhi, Delhi, India. He is currently a Senior Scientist at the Defence Geoinformatics Research Establishment (DGRE), Defence Research and Development Organisation (DRDO), New Delhi, India. His research interests span geospatial intelligence, applied machine learning, and remote sensing. He provided research direction and conceptual guidance during the design and evaluation phases of this work. 
\end{IEEEbiography}

\begin{IEEEbiography}[{\includegraphics[width=1in,height=1.25in,clip,keepaspectratio]{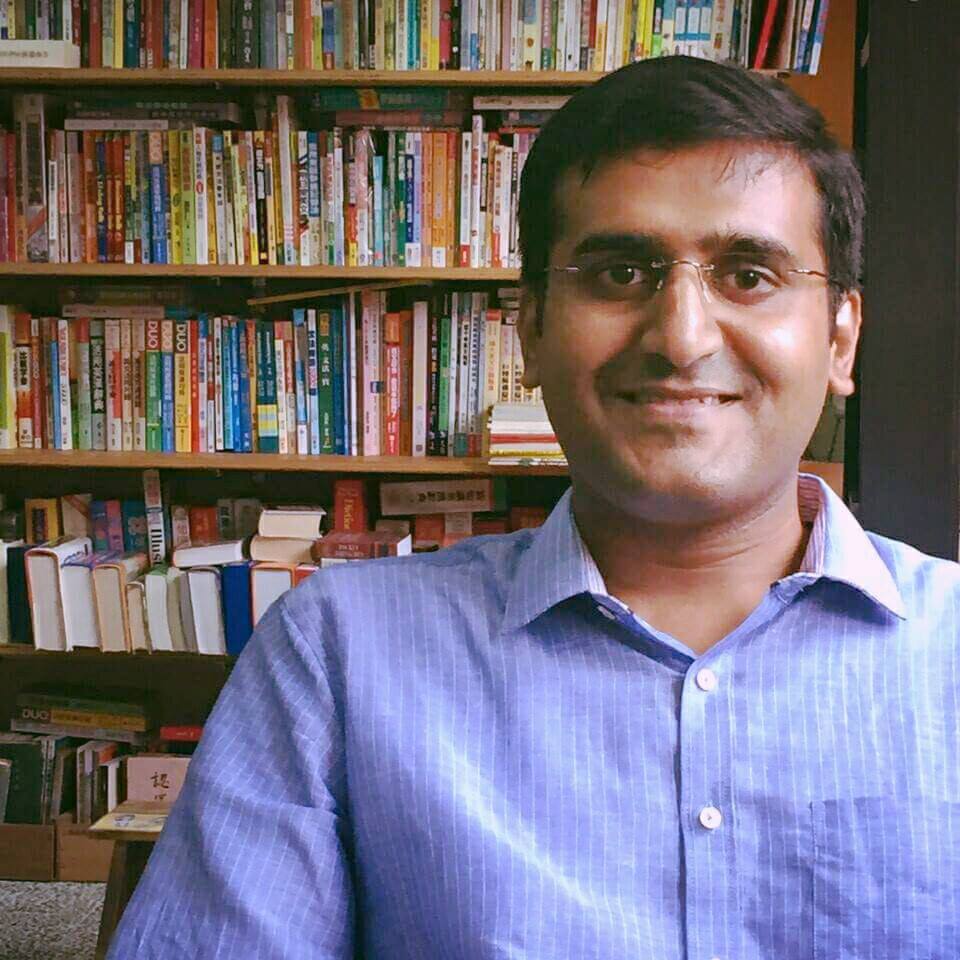}}]{Dr. Kathiravan Srinivasan}
(Senior Member, IEEE) received the Ph.D.\ degree from Anna University, Chennai, India. He is currently a Professor with the School of Computer Science and Engineering, Vellore Institute of Technology, Vellore, India. His research interests include machine learning, deep learning, computer vision, and continual learning systems. He supervised the research trajectory of this work, provided computational infrastructure, and contributed to experimental design, result interpretation, and manuscript revision. Correspondence regarding this article should be addressed to Kathiravan Srinivasan (e-mail: kathiravan.srinivasan@vit.ac.in).
\end{IEEEbiography}

\end{document}